\documentclass{article}

\PassOptionsToPackage{numbers, sort&compress}{natbib}

\usepackage{NeurIPS_2023}

\usepackage[utf8]{inputenc} % allow utf-8 input
\usepackage[T1]{fontenc}    % use 8-bit T1 fonts
\usepackage{url}            % simple URL typesetting
\usepackage{booktabs}       % professional-quality tables
\usepackage{nicefrac}       % compact symbols for 1/2, etc.
\usepackage{microtype}      % microtypography
\usepackage[utf8]{inputenc}
\usepackage{moreverb}
\usepackage{mathtools}
\usepackage{amsmath}
\usepackage{amssymb}
\usepackage{algorithmic}
\usepackage[linesnumbered,lined,ruled]{algorithm2e}
\usepackage{graphics}
\usepackage{graphicx}
\usepackage{subfigure}
\usepackage{caption}
\usepackage{enumitem}
\usepackage{extarrows}
\usepackage{color}
\usepackage{framed}
\usepackage{wrapfig}
\usepackage{bm}
\usepackage{mathrsfs}
\usepackage{mathabx}
\usepackage{multirow}
\usepackage{longtable}
\usepackage{hyperref}
\usepackage{paralist}
\usepackage{indentfirst}
\usepackage{relsize}
\usepackage{extarrows}
\usepackage{lineno,xcolor}
\usepackage{upgreek}
\usepackage{bm}
\usepackage{mwe}
\usepackage{xcolor}
\usepackage[flushleft]{threeparttable}
\usepackage[normalem]{ulem}
\usepackage{caption}
\usepackage{rotating} 

\hypersetup{
bookmarks=true,
bookmarksopen=true,
bookmarksnumbered=true,
unicode=false,
pdftoolbar=true,
pdfmenubar=true,
pdffitwindow=false,
pdfstartview={FitH},
pdftitle={My title},
pdfauthor={Author},
pdfsubject={Subject},
pdfcreator={Creator},
pdfproducer={Producer},
pdfkeywords={keywords},
pdfnewwindow=true,
colorlinks=true,
linkcolor=blue,
citecolor=blue,
filecolor=magenta,
urlcolor=blue
}

\title{TLNets: Transformation Learning Networks for long-range time-series prediction}

\author{Wei Wang$^1$, Yang Liu$^2$, Hao Sun$^{1,}$\thanks{Corresponding author}  \vspace{0pt} \\
{\small $^1$Gaoling School of Artificial Intelligence, Renmin University of China, Beijing, China;} \\ {\small $^2$School of Engineering Science, University of Chinese Academy of Sciences, Beijing, China;} \vspace{0pt}  \\
{\small Emails: \url{xiaokeaiww888@yeah.net};~~\url{liuyang22@ucas.ac.cn};~~\url{haosun@ruc.edu.cn}}
}

\begin{document}
\maketitle

\begin{abstract}
Time series prediction is a prevalent issue across various disciplines, such as meteorology, traffic surveillance, investment, and energy production and consumption. Many statistical and machine-learning strategies have been developed to tackle this problem. However, these approaches either lack explainability or exhibit less satisfactory performance when the prediction horizon increases. To this end, we propose a novel plan for the designing of networks' architecture based on transformations, possessing the potential to achieve an enhanced receptive field in learning which brings benefits to fuse features across scales. In this context, we introduce four different transformation mechanisms as bases to construct the learning model including Fourier Transform (FT), Singular Value Decomposition (SVD), matrix multiplication and Conv block. Hence, we develop four learning models based on the above building blocks, namely, FT-Matrix, FT-SVD, FT-Conv, and Conv-SVD. Note that the FT and SVD blocks are capable of learning global information, while the Conv blocks focus on learning local information. The matrix block is sparsely designed to learn both global and local information simultaneously. The above Transformation Learning Networks (TLNets) have been extensively tested and compared with multiple baseline models based on several real-world datasets and showed clear potential in long-range time-series forecasting. 
\end{abstract}

\section{Introduction}
Time-series prediction is a crucial problem that is commonly encountered in many disciplines, which has numerous applications, such as weather forecasting, traffic prediction, stock market analysis, and electricity consumption forecasting. Over the years, many statistical and analytical methods have been developed to tackle this issue. Earlier, researchers used statistical measures like mean and variance to devise models such as ARMA \cite{mcleod1983diagnostic} and ARIMA \cite{ho1998use, zhang2003time}. These kinds of models have been characterized by highly targeted and good robustness. However, they are featured poor generalization ability and weak performance in multivariate time-series prediction.

Recently, many deep learning models have been developed to tackle time-series forecasting problems, e.g., Recurrent Neural Networks (RNN) \cite{Bouktif2018OptimalDL, lai2018modeling, bahdanau2014neural, hewamalage2021recurrent}, Temporal Conv-based models \cite{bai2018empirical, vorbach2021causal, aksan2019stcn, luo2019conv, hewage2020temporal,lara2020temporal, wan2019multivariate, liu2021time}, Transformer models \cite{kitaev2020reformer, zhang2018visual, wu2021autoformer, shen2022tcct, madhusudhanan2021yformer,2020arXiv201207436Z, TianZhou2022FEDformerFE}, and Linear-type model \cite{AilingZeng2022AreTE}. The nature of such an issue is supervised learning in which inputs are the collected sequence data and outputs are the predicted sequences at future time steps. The models aim to project the temporal information of the time series at previous steps into the future horizon for prediction, via capturing correlation across multivariate sequences.

%Recently, many deep learning models \cite{kourentzes2014neural, rahman2015layered, torres2018scalable, kitaev2020reformer, zhang2018visual, wu2021autoformer, shen2022tcct, madhusudhanan2021yformer,2020arXiv201207436Z,TianZhou2022FEDformerFE} have been developed to tackle time-series forecasting problems. Such problems typically involve supervised learning wherein the collected sequence data form the inputs and the predicted sequences at future time steps form the outputs. These models aim to project temporal information from previous time steps into the future horizon for accurate prediction while capturing correlations across multivariate sequences.

Despite their efficacy, these methods either lack explainability or exhibit less satisfactory performance when the prediction horizon dramatically increases. Generally speaking, the above-mentioned deep learning methods are built upon the process of receptive field learning (RFL), which maps features from small/local receptive fields (e.g., via Conv, attention) to big/global ones (e.g., via deep layers). Balanced local and global RFL in theory brings benefits for better representation learning. While previous work by Luo \textit{et al.} \cite{Luo_Li_Urtasun_Zemel_2016} attempted to analyze models using the effective receptive field, a comprehensive definition and demonstration of RFL are missing.

We define any model, which learns through weights-multiplied features (usually adjacent) as new features followed by summation or multiplication of them, can be considered an example of RFL. Instead of achieving big receptive fields via employing deep layers, we hypothesize that transformation of the features into a \textit{properly defined domain} (e.g., Fourier domain, orthonormal domain) that naturally possesses big receptive fields is an alternative. Hence, we introduce four generalized transformation mechanisms as bases to construct the learning model including the Fourier Transform (FT), Singular Value Decomposition (SVD), matrix multiplication, and Conv block. We develop four learning models based on the above building blocks, namely, FT-Matrix, FT-SVD, FT-Conv, and Conv-SVD, all based on our RFL definition and proofs on the linking between these four blocks. Note that the FT and SVD blocks are capable of learning global information, while the Conv blocks focus on learning local information. The matrix block is sparsely designed to learn both global and local information simultaneously. We believe RFL has the potential to provide a more interpretable framework for understanding increasingly complex models and guiding network design.

%According to the relationship between FT, SVD, Conv, and matrix multiplication, we further demonstrate that convolutional neural networks can be represented as ordered matrix multiplication from front to back (unless no activation function is present, in which case almost any convolutional model can be simply written as matrix multiplication), with middle matrices responsible for learning data patterns based on varying receptive fields. This approach enables us to write any model in a specific formula and identify differences between different networks, and determine which parts contribute to making a particular model surpass other models. RFL has the potential to provide a more interpretable framework for understanding increasingly complex models and guiding network design.

%Furthermore, in \textcolor{blue}{S.I. Appendix D}, we demonstrate that the basic structure of the Transformer, multi-head attention, can also be represented as a form of matrix multiplication.

The contributions of our paper are summarized as follows:

\begin{itemize}
\item We introduced a general definition of RFL in deep learning, based on which we propose several transformation learning newtorks (TLNets), namely, FT-Matrix, FT-SVD, FT-Conv, and Conv-SVD, to achieve balanced local and global RFL. We showed that learning in a transformed feature domain achieves better performance. \vspace{3pt}

%We give a definition of RFL in deep learning. All approaches in deep learning that use parameterized learning of multiple features (often involving multiplication of multiple features with corresponding parameters) and combine abstracted features through addition or multiplication can be considered RFL. Just like human, deep learning also rely on global and local information viewed by networks. Moreover, all the mathematical transformations that fulfil these requirements could be used in deep learning. Conditioned on that we propose a standard paradigm for neural networks. \vspace{3pt}

%\item Building on the standard paradigm, we further introduce four newly-developed deep learning architectures, namely FT-Matrix, FT-SVD, FT-Conv, and Conv-SVD. \vspace{3pt}
% that projects feature learning into the orthogonal spaces via SVD and Fourier spaces via Fourier transform~(FT). The network architecture consists of (1) the Fourier neural operator block that learns the dominant frequency contents that reflect the time-series variation at both the short and the long horizon, and (2) the Singular Value Decomposition~(SVD) block that captures the correlation between multiple channels of the time series.

\item We demonstrated the relationship between Conv, Fourier Transform (FT), and matrix multiplication, along with their corresponding receptive field information, and explained why typical neural networks such as CNN require deep layers to maintain big RFL. \vspace{3pt}

%. Based on these relationships, we explained why convolutional neural networks require deep architectures, explored the role of activation functions in convolution, and showed through mathematical proofs that CNNs learn low-frequency features in shallow layers and high-frequency features in deeper ones. Additionally, we showcased the theoretical advantages of using Fourier neural operators for deep learning compared to standard convolutional networks. \vspace{3pt}
%Building on this foundation, we analyzed the characteristics and limitations of convolutional networks, proposed the receptive field theory of deep learning (such as local/small fields versus global/big fields), and presented the general framework of deep learning.

\item We extensively tested and compared our proposed models with multiple baseline models, using several real-world datasets. Results demonstrate that our proposed models outperform the existing methods.
\end{itemize}

\section{Related Work}
\label{section:Related Work}

\textbf{RNN}: Recurrent neural networks (RNN) were popular in time-series forecasting a few years ago, emphasizing the importance of sequential dependency. RNNs consist of various gated units to learn the connection between sequence positions \cite{Bouktif2018OptimalDL, lai2018modeling, bahdanau2014neural, hewamalage2021recurrent}. The basis of RNN is the Markov Chain process in mathematics. However, gradient vanishing, large training efforts, and fast error accumulation across the temporal horizon remain key bottlenecks.

\textbf{CNN}: The Temporal Convolutional Network (TCN) could serve as another alternative solution for time-series foresting \cite{bai2018empirical, vorbach2021causal, aksan2019stcn, luo2019conv, hewage2020temporal}, which was promoted by Wavenet autoregressive model \cite{oord2016wavenet}. It outlines causal convolution to avoid watching future data. Besides, in order to capture the long-term information in time series, it employs dilated convolution. Some other similar models were also developed \cite{lara2020temporal, wan2019multivariate, liu2021time}, where the most effective one is the state-of-the-art SCINet \cite{liu2021time} which secures good results on both long-range and short-term time series forecasting compared with other existing Conv-based models. 

\textbf{Transformer}: Transformers have almost dominated deep learning and show critical potential in solving time-series forecasting problems. The multi-head attention architecture can extract information, and the position embedding can retain sequence position information~\cite{kitaev2020reformer, zhang2018visual, wu2021autoformer, shen2022tcct, madhusudhanan2021yformer}. However, the computational complexity of Transformers is high, and setting hyperparameters has a considerable impact on models that use Transformers as a backbone. To address this, the Informer, Autoformer, and Fedformer models~\cite{2020arXiv201207436Z, wu2021autoformer, TianZhou2022FEDformerFE} were developed. Although effective for long-sequence forecasting, model performance deteriorated substantially as the prediction horizon increased.

\textbf{Linear}: Early works on time-series forecasting employed fully connected neural networks \cite{kourentzes2014neural, rahman2015layered, torres2018scalable}. However, these networks failed to learn the sequential/temporal dependency of time series effectively. Recently, \cite{AilingZeng2022AreTE} solved those problems and proposed that the Transformer is not the best solution for time-series forecasting. They instead used linear methods to achieve state-of-the-art results on both long-range and short-term time series forecasting, compared with existing models.

\section{Preliminary}
\label{section:Preliminary}

First of all, we give a conceptual description of time-series prediction. Let $\mathbf{X}$ be a time-series sequence. Our purpose is to use the sequences at previous $T$ time steps, i.e., $\mathbf{X}_{t-T+1: t}=\left\{\mathbf{x}_{t-T+1}, \ldots, \mathbf{x}_{t}\right\}$ to predict the sequences at future $\tau$ steps, i.e., $\mathbf{X}_{t+1: t+\tau}=\left\{\mathbf{x}_{t+1}, \ldots, \mathbf{x}_{t+\tau}\right\}$. Here, $\mathbf{x}_{t}\in R^d$ represents the time-series sequences at time $t$, where $d$ denotes the dimension of the sequences~(note that $d>1$ denotes multivariate sequences). We seek to develop models to forecast $\mathbf{X}_{t+1: t+\tau}$ given $\tau$ is large which represents a long-range horizon. This is essentially a supervised learning problem -- establishing a neural operator that projects the temporal information of the time series at previous steps into the future horizon for prediction.

Li \textit{et al.} \cite{2020arXiv201008895L} utilized Fourier Transform and introduced a new paradigm for neural networks called the Fourier Neural Operator (FNO). The FNO model is defined as follows:
\begin{equation}
\begin{aligned}
v_{t+1}(x):=\sigma\left(W v_t(x)+\mathcal{F}^{-1}\left(R_\phi \cdot\left(\mathcal{F} v_t\right)\right)(x)\right), \quad \forall x \in D,
\end{aligned}
\end{equation}
where $W: \mathbb{R}^{d_v} \rightarrow \mathbb{R}^{d_v}$ is a linear transformation, and $\sigma: \mathbb{R} \rightarrow \mathbb{R}$ is an activation function. $\mathcal{F}$ and $\mathcal{F}^{-1}$ are the FT and its inverse; $R_{\phi}: \mathbb{Z}^{d} \times \mathbb{R}^{d_{v}} \rightarrow \mathbb{R}^{d_{v} \times d_{v}}$ denote the parameters learned from data. $v_t$ and $v_{t+1}$ represent the input and output, respectively. The introduction of FNO represents a significant advancement in deep learning as it provides a standard paradigm for describing neural networks. However, the paper lacks a clear theoretical explanation for this paradigm, and it is not universally applicable as it is only suitable for the networks presented in that paper.

\begin{figure}[t!]
  \centering
  \includegraphics[width=0.95\textwidth]{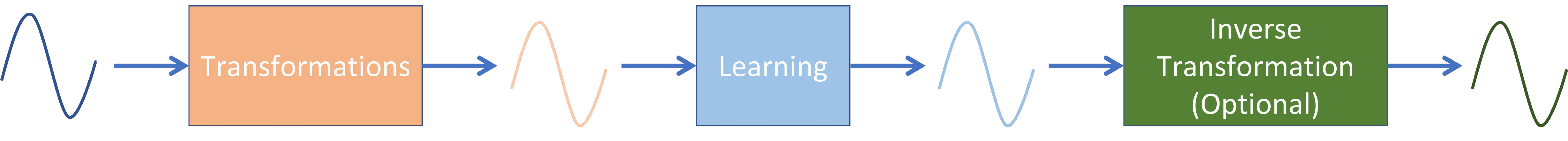}
\caption{Schematic of deep learning that involves transforming input features into a latent space/domain for learning with big receptive fields. Note that inverse transformation is optional, depending on the specific transformation method used.}   
\label{fig:trans}
\end{figure}

Based on the definition of RFL, we propose a general paradigm for neural networks. We hypothesize that any transformation capable of gaining a receptive field (e.g., CNN, FT, wavelet transformation, Transformer, SVD, etc.) can be utilized in deep learning. Drawing inspiration from this concept, we suggest that the neural operator can be rewritten as:
\begin{equation}
\begin{aligned}
v_{t+1}(x) 
= &~W_Sv_{t}(x)  + H^{-1} [W_H H [v_{t}(x)]] + \sigma \big(K^{-1} [W_K K [v_{t}(x)]] \\ &+  G^{-1} [W_G \odot G [v_{t}(x)]],
...\big)
\end{aligned}
\label{eq:paradigm}
\end{equation}
where $K$, $G$ and $H$ denote mathematical transformations, such as functional or matrix transformation~(the receptive field of transformations are required to change according to the task), and $K^{-1}$, $G^{-1}$ and $H^{-1}$ are the inverse. $W_K$, $W_G$ and $W_H$ are latent parameters to be learned from data. $W_S$ is a designed sparse matrix. By choosing the appropriate transformation, we could achieve flexible receptive field learning. The schematic process could be seen in Figure \ref{fig:trans}. In this paper, we use FT, Convolution, SVD and matrix multiplication as bases for transformation learning.

\section{Design of TLNets}
\label{section: Networks Design}
According to Eq \eqref{eq:paradigm}, we design four networks based on FT, sparse matrix multiplication, SVD, and Conv. Firstly, we introduce the sparse matrix block, FT block, and SVD block. Then, we will provide detailed specifications for implementing designing the four networks based on these building blocks.

\subsection{Sparse Matrix Block}
\label{section: Sparse Matrix Block}

\subsubsection{The Relationship between Convolution and Sparse Matrix Multiplication}
\label{section: The Relationship between Convolution and Matrix Multiplication}

Convolution has become the dominant learning operator in deep neural networks thanks to its proven effectiveness across a wide range of tasks. However, explaining how convolution works within the context of deep learning is non-trivial and mathematically challenging since convolution is not an explicit calculation involving simple addition and Hadamard products. To this end, we represent convolution in the form of matrix multiplication for straightforward interpretability.

%poses significant challenges since convolution is not an explicit calculation involving simple addition and Hadamard product

In \textcolor{blue}{S.I. Appendix B}, we provide the proof of 1D Conv expressed as matrix multiplication. We refer to the sparse matrix turned from the Conv kernel as the Conv matrix. This approach allows us to gain a more comprehensive understanding of the entire process of a 1D CNN, which can be expressed as:
\begin{equation}
\begin{aligned}
\mathbf{y} =
\sigma\{\mathbf{h}_n...\sigma[\mathbf{h}_1\sigma(\mathbf{h}_1\mathbf{X})]\}.
\end{aligned}
\end{equation}
Here, $\mathbf{h}_1...\mathbf{h}_n$ represent the Conv matrices used in the network. Importantly, it should be noted that these sparse matrices are designed according to Conv, with most of their elements being zero. The non-zero elements are determined by factors such as kernel sizes, input channels, and output channels. $\mathbf{X}$ denotes the input to the model, while $\mathbf{y}$ refers to the model's output. The entire process can be understood as an ordered matrix multiplication. Computation follows a fixed order from front to back due to the inclusion of activation functions, which necessitates this specific ordering of computation.

We can view the learning process of convolution as RFL, where the Conv matrices are sparse and learn patterns defined by Conv. Based on this concept, we can design the sparse matrix ourselves. One reason for the popularity of convolution is attributed to its ability to reduce computation. According to the shape of the sparse matrix formed from convolution, we know that it fulfils a large number of zero parameters, which could lead to computational overhead. Convolution effectively avoids these parameters and has therefore become widely adopted. Another reason is that learning a dense matrix is a challenging task.

\subsubsection{The Designment of Sparse Trainable Matrix}
We have proved that Conv can be written as matrix multiplication. The single-layer Conv has a notable drawback, which is a small receptive field (See \textcolor{blue}{S.I. Appendix C.1}). Typically, the Conv size takes a small value (e.g., 3, 5), so the Conv matrix consists mostly of zeros. As is widely known, as neural networks become deeper, their receptive fields tend to increase in size. This characteristic can be observed through changes in the Conv matrix, which we discuss in greater detail in \textcolor{blue}{S.I. Appendix C.2}. We found that the receptive field becomes larger and has a regular pattern, which is discerned from the Conv matrix in \textcolor{blue}{S.I. Appendix C.2}. 

Secondly, our goal is to identify the targeted Conv matrix $\mathbf{h}$. However, a problem arises because all features share the same pattern, with only the combination of convolutions. In order to overcome this constraint, existing CNN models utilize activation functions. These functions break the pattern constraints of the kernels. Additionally, shallow networks with small kernel sizes have a limited receptive field, making it difficult to perform tasks using shallow networks. To address this issue, continuous convolutions are introduced to optimize $\mathbf{h}$, e.g., expressed as $\sigma[\mathbf{x} \circledast \mathbf{h}_0]\}\circledast \mathbf{h}_1$. This produces a two-layer Conv network with an activation function and a larger receptive field for each point on the feature map in the second layer. However, the first layer's features always have the same small Conv kernel, which means that all data use the same pattern. The activation function plays an important role in breaking the shared pattern across all data in all layers. This explains why traditional networks require activation functions and deeper network depths. Activation functions are nonlinear operations that can break the pattern constraints of the kernels, and deeper layers result in a larger receptive field. However, if we do not restrict the size of $\mathbf{h}$, we can eliminate numerous convolutional layers altogether.

\begin{wrapfigure}[16]{r}{0.5\textwidth}
  \centering
  \vspace{-12pt}
\includegraphics[width=0.99\linewidth]{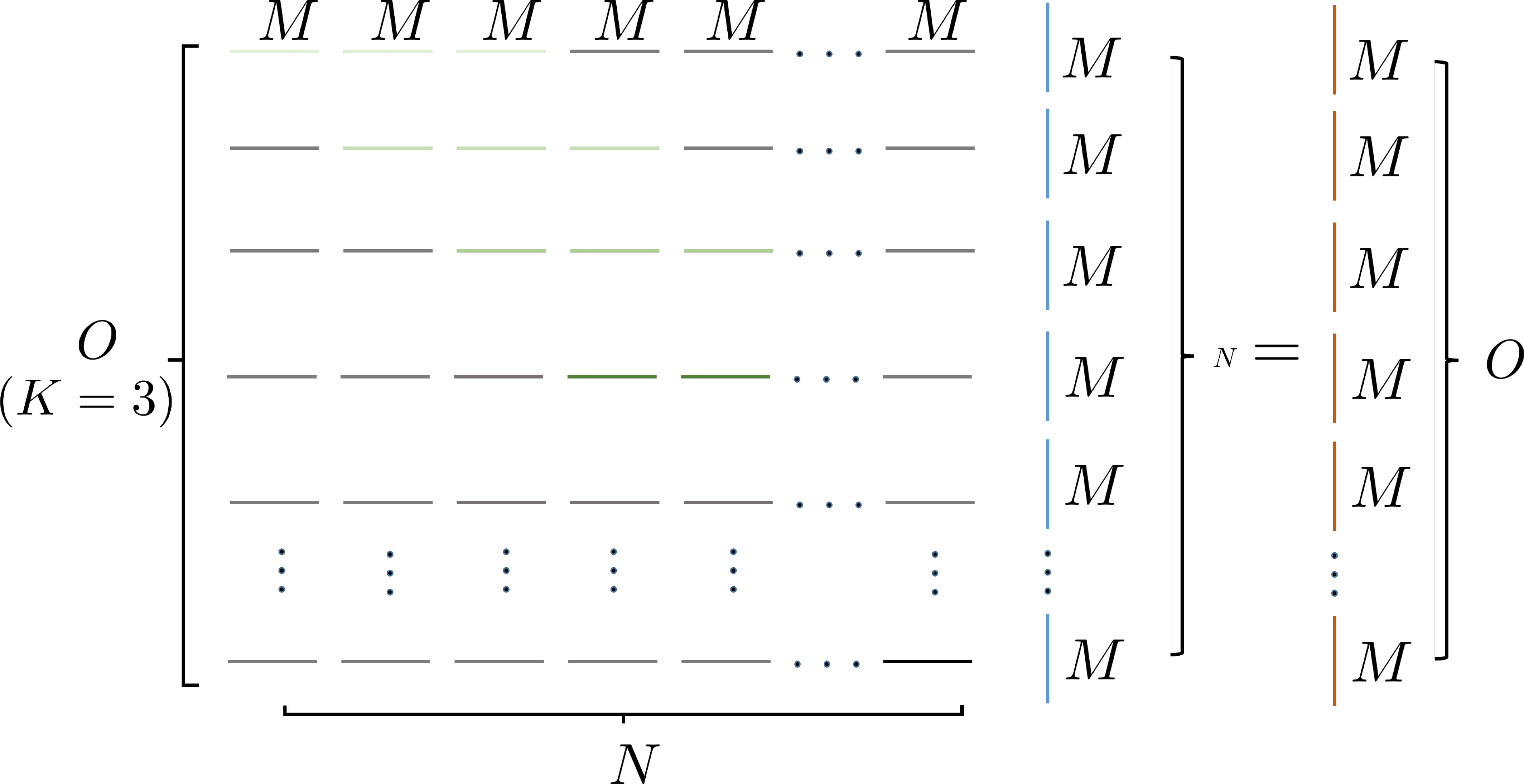}
\caption{The sparse trainable matrix. The grey lines stand for zeros. The different blue lines mean the parameters with different parents. $M$ is the number of features. $N$ is the input channel. $O$ is the number of output channels. $K$ is the number of kernels.}     
\label{fig:Sparse_Matrix}
\end{wrapfigure}

To address the issues of receptive field learning and pattern recognition, we propose utilizing a sparse trainable matrix as a solution, as illustrated in Figure \ref{fig:Sparse_Matrix}. Initially, we generate a random matrix based on the input and output shapes, which is then optimized through forward and backward propagation, known as a parameter matrix. Subsequently, we construct a matrix filled with zeros and ones, referred to as a shape matrix, whose specific shape is determined by researchers. During forward propagation, we calculate the Hadamard product between the parameter and shape matrices, allowing us to disregard the effects of irrelevant parameters and features on the results. Through this technique, we can incorporate various patterns and expand our receptive field. Figure \ref{fig:Sparse_Matrix} showcases the structure of the sparse matrix. By implementing this matrix, we can simultaneously utilize kernel patterns of varying sizes.

\subsection{FT Block}
\label{section: FT Block}
\subsubsection{The Relationship between Convolution and FT}
\label{section: The Relationship between Convolution and FT}
The frequencies learned by a model are crucial in deep learning. Some studies have indicated that neural networks have difficulty learning high-frequency components in shallow layers~\cite{ZhiQinJohnXu2019FrequencyPF, ZhiQinJohnXu2019TrainingBO, LingTang2022DefectsOC}, but they cannot provide definitive proof. To address this issue, we investigate the relationship between kernel size and frequencies in the Fourier domain and prove that the frequencies learned by convolution are determined by the kernel size. According to the convolution theory $\mathbf{x} \circledast \mathbf{h} = \mathbf{F}^{-1} (\mathbf{F} \mathbf{x} \odot \mathbf{F} \mathbf{h})$ (The proof could be found in \textcolor{blue}{S.I. Appendix A}). The frequencies can be learned by a convolution with a kernel size of three as shown in \textcolor{blue}{S.I. Appendix C.3}. The shallow layers only contain low-frequency information, as most of the values in $\mathbf{h}$ are zero. Therefore, the highest frequency that can be learned from convolution is fixed when we define the kernel size. If we set the kernel size to three, only a portion of the low-frequency information can be obtained from Conv.

The changes in frequencies resulting from two-layer convolution are demonstrated as follows
\begin{equation}
(\mathbf{x} \circledast \mathbf{h}) \circledast \mathbf{h} = \mathbf{F}^{-1}\{\mathbf{F}[\mathbf{F}^{-1} (\mathbf{F} \mathbf{x} \odot \mathbf{F} \mathbf{h})]\odot \mathbf{F} \mathbf{h}\} = \mathbf{F}^{-1}\ [ (\mathbf{F} \mathbf{x} \odot \mathbf{F} \mathbf{h})\odot \mathbf{F} \mathbf{h}].
\label{Eq: general_concolution_two}
\end{equation}
As we are focused on illustrating the general trend of learnable frequencies, we have ignored the activation function in the above equation. Adding an activation function would only scale the output of the convolution, and may potentially decrease the highest frequencies by setting some elements to zero or not changing the frequencies. However, it is evident from Eq. \eqref{Eq: general_concolution_two} that the frequencies that can be learned increase with deeper network architectures. Hence, the lower layers of networks learn low-frequency information, while the higher layers gradually acquire high-frequency information.

\subsubsection{The Designment of FT}
\label{section: The Designment of FT}

\begin{wrapfigure}[12]{r}{0.35\textwidth}
  \centering
  \vspace{-30pt}
  \includegraphics[width=0.99\linewidth]{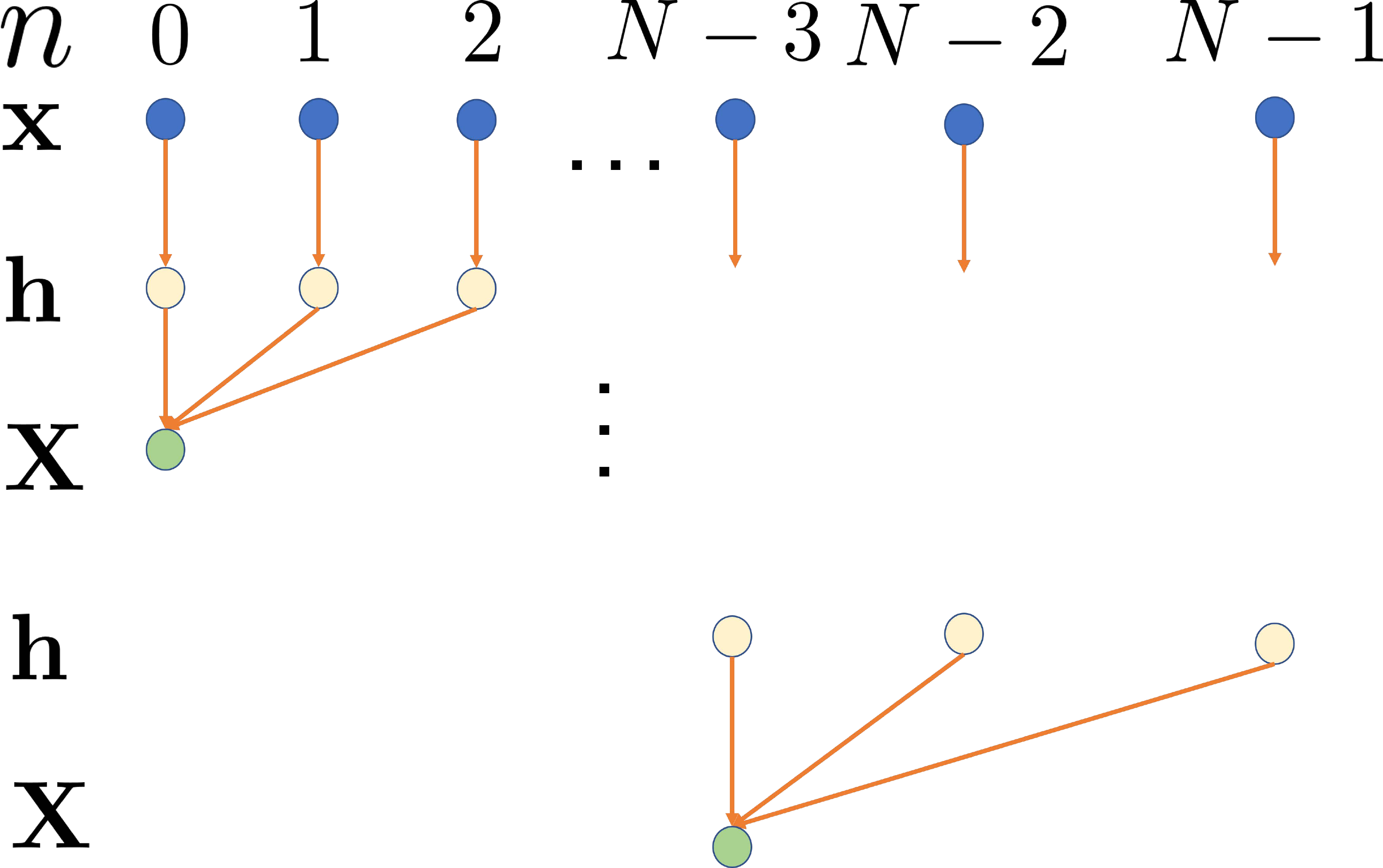}
\caption{The convolution operation on the sequence. $n$ is the serial number. $\mathbf{x}$ is a sequence. $\mathbf{h}$ is the convolutional kernel and the kernel size is three. $\mathbf{X}$ is the output after convulution. }   
\label{fig:Conv_series}
\end{wrapfigure}

\begin{wrapfigure}[20]{r}{0.63\textwidth}
  \centering
\includegraphics[width=0.99\linewidth]{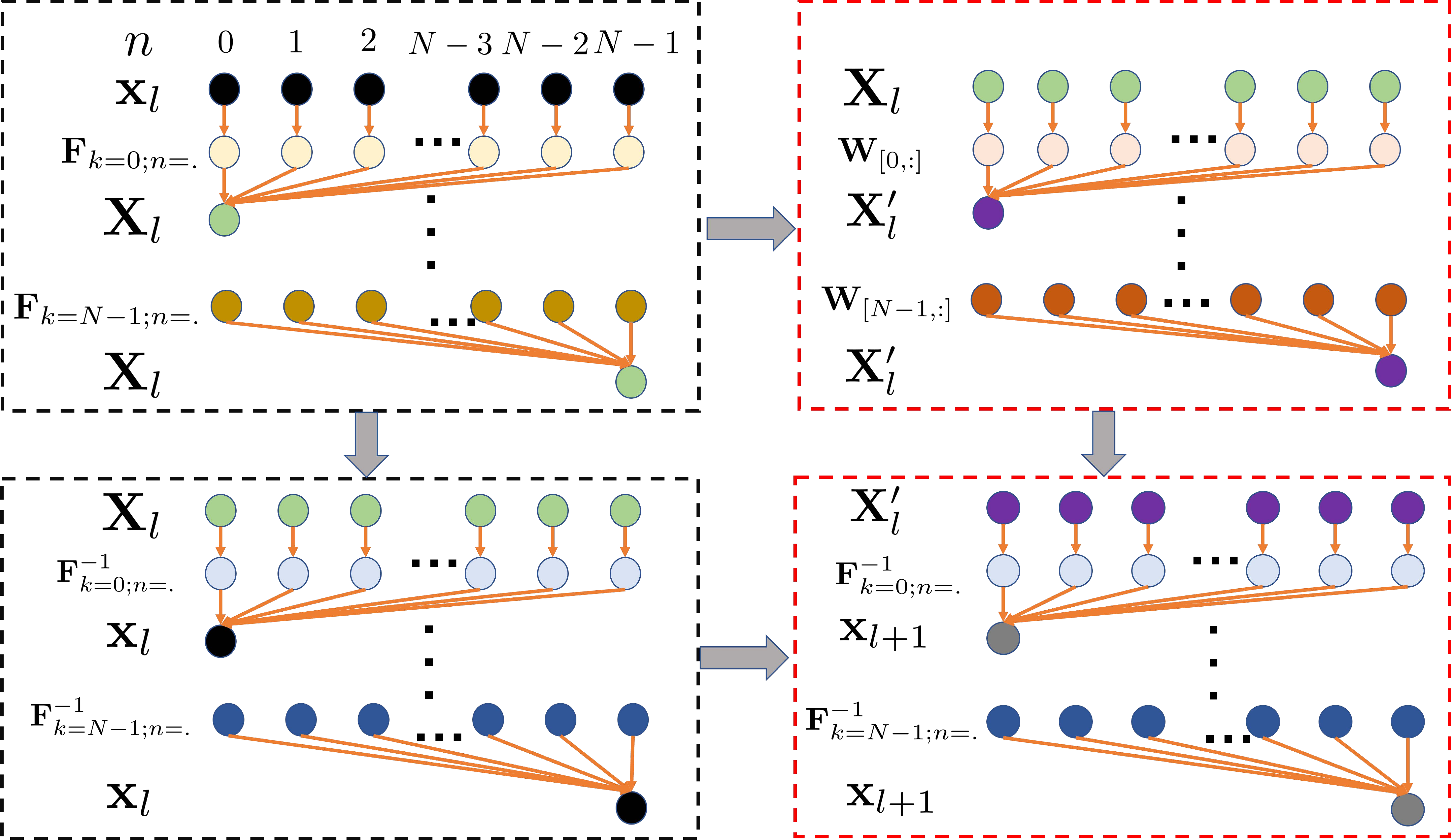}
\caption{The left part is the FT on a latent sequence. $n$ is the serial number. $\mathbf{x}_l$ and $\mathbf{x}_{l+1}$ are latent sequences. $\mathbf{F}$ is the FT matrix and $\mathbf{F}^{-1}$ is inverse matrix of the FT. The right part is the way of learning the change of sequence in the Fourier domain. $\mathbf{W}$ is the weight matrix. $\mathbf{X}_l$ is the FT output. $\mathbf{X}_l'$ is the latent features learned in the Fourier domain.}     
\label{fig:FT_series}
\end{wrapfigure}

We know that the relationship between convolution and FT can be expressed as $\mathbf{x} \circledast \mathbf{h}=\mathbf{F}^{-1} [ \mathbf{F} \mathbf{x} \odot \mathbf{F} \mathbf{h}]$. 
Suppose that $\mathbf{h}$ is the target, but it is difficult to obtain. However, we can still learn $\mathbf{h}$ in the frequency domain, namely, $\mathbf{x} \circledast \mathbf{h}=\mathbf{F}^{-1} \mathbf{W} \mathbf{F} \mathbf{x}$. Thus, we could design an upgrade, $\mathbf{F}^{-1} \mathbf{W} \mathbf{F} \mathbf{x}$, of traditional convolution in neural networks. Based on this, we can learn features in the frequency domain and achieve a global receptive field. This can be clearly seen through Figures \ref{fig:FT_series} and \ref{fig:Conv_series}. As shown, compared with convolution, the red box in Figure \ref{fig:FT_series} has a larger receptive field. Each point in the Fourier domain can collect all information in the original domain.

\subsection{SVD Block}
\label{section: SVD Block}

Encouraged by the application of FT  in deep learning and RFL, we consider adopting other modalities of functional transformation or matrix decomposition. In this paper, we introduce the SVD learning block. The formula of SVD can be written as:
\begin{equation}
\texttt{SVD}(\mathbf{x}) = \mathbf{U} \mathbf{S} \mathbf{V} 
\label{SVD_trans}
\end{equation}
where $\mathbf{U}$ and $\mathbf{V}$ are the orthonormal eigenvector matrices, $\mathbf{S}$ is the singular value matrix. Firstly, we conduct SVD on both the $l$th layer's input $\mathbf{x}_l = \mathbf{U}_\mathbf{x} \mathbf{S}_\mathbf{x} \mathbf{V}_\mathbf{x}$ and the trainable weight $\mathbf{\Phi}_l = \mathbf{U}_\mathbf{\Phi} \mathbf{S}_\mathbf{\Phi} \mathbf{V}_\mathbf{\Phi}$. Then, we compute the Hadamard product between $\{\mathbf{U}_\mathbf{x}, \mathbf{S}_\mathbf{x}, \mathbf{V}_\mathbf{x}\}$ and $\{\mathbf{U}_\mathbf{\Phi}, \mathbf{S}_\mathbf{\Phi}, \mathbf{V}_\mathbf{\Phi}\}$, correspondingly. By calculating the matrices of the Hadamard product results, we will get the output of the SVD Block. 

The reason why we introduce SVD can be summarized as follows. The purpose of machine learning is to establish a parametric model which is trained against given data to solve the target problem. Some researchers showed that orthogonal parameters are helpful for achieving better model convergence \cite{2020arXiv200405867H}. While there is a critical problem strictly maintaining the parameter orthogonality during the backpropagation-based model training process is challenging. Given the strict orthogonal property of 
SVD, our method can guarantee the decomposed weights~(i.e., $\{\mathbf{U}_\mathbf{\Phi}, \mathbf{S}_\mathbf{\Phi}, \mathbf{V}_\mathbf{\Phi}\}$) are orthogonal all the time. Besides, since $\{\mathbf{U}_\mathbf{x}, \mathbf{S}_\mathbf{x}, \mathbf{V}_\mathbf{x}\}$ are from the same input vector $\mathbf{x}_l$, there are intimate relationships between them. If we generate the weights randomly and perform the product on the decomposed input, the weights will not fulfil the same pattern in theory. Therefore, we propose to conduct SVD on the trainable weights to enable the update of the decomposed input, which retains orthogonality for better reconstruction of the targeted time series.

Another reason is that we know the rows' direction of the data reflects the data's change with time, while the columns' direction reflects the data's change with features. Suppose, the size of $\mathbf{x}, \mathbf{U}, \mathbf{S}, \mathbf{V}$ are $(m,n)$, $(m,m)$, $(m,m)$ and $(m,n)$.  So we could rewrite the SVD decomposition as:
\begin{equation}
\texttt{SVD}(\mathbf{x})=\Sigma_{i=0}^m \mathbf{U}_{[:,i]} \mathbf{S}_{[i,i]} \mathbf{V}_{[i,:]} 
\label{SVD_trans_sigma}
\end{equation}
where $\mathbf{U}_{[:,i]}$ and $\mathbf{V}_{[i,:]}$ are the eigenvectors with respect to eigenvalue $\mathbf{S}_{[i,i]}$ in the time and feature directions. The scales of eigenvalues show the importance of eigenvectors. Hence, SVD could predict the time series from the eigen/modal perspective and essentially captures spatiotemporal dynamics for multivariate time-series forecasting.

\subsection{TLNets for Time-Series Forecasting}
\label{section: Networks for Time-Series Forecasting}

\begin{figure*}[t!]
\centering
\includegraphics[width=0.9\textwidth]{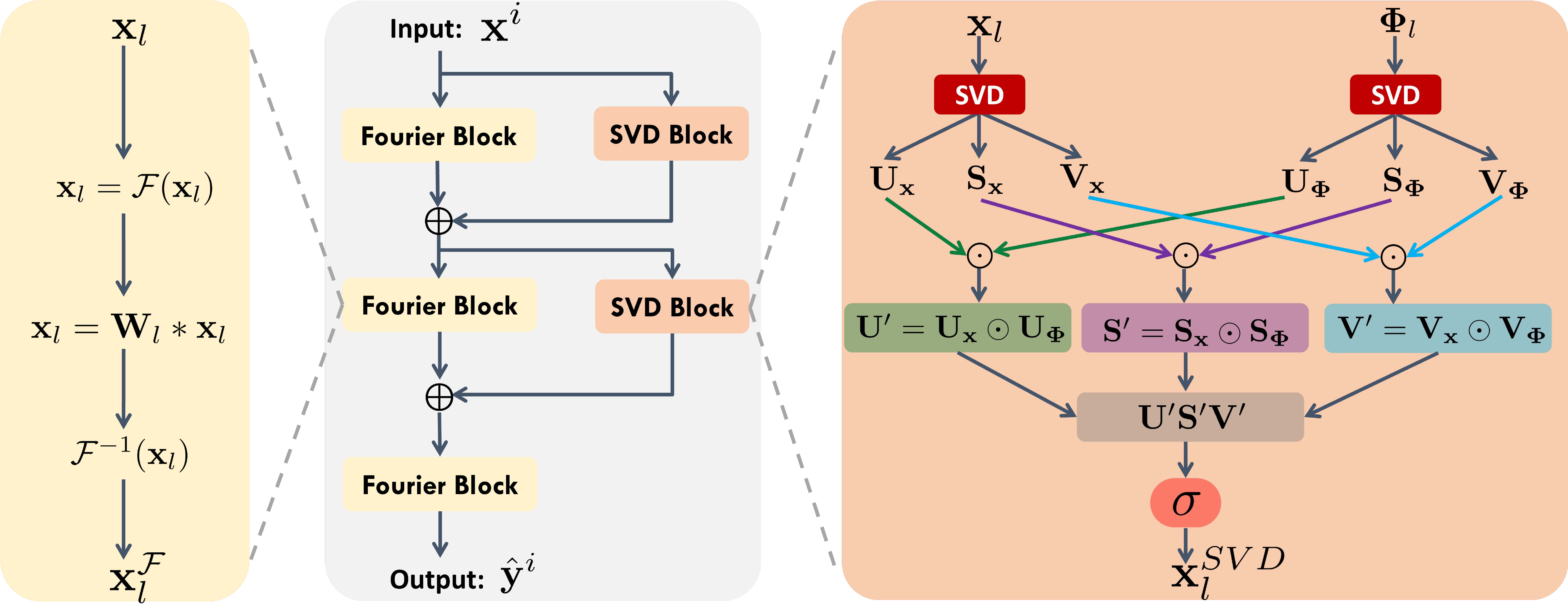}
\caption{The schematic architecture of FT-SVD which shows the basic operating units of the proposed neural network. The left and right parts are the details of the Fourier block and SVD block.}     
\label{fig:FFT_network}
\end{figure*}

Using the fundamentals of the theory described above, we have developed four new architectures of TLNets. One of them is FT-SVD, as shown in Figure \ref{fig:FFT_network}. The network architecture comprises (i) the Fourier neural operator block, which learns the dominant frequency contents reflecting the time-series variation at both short and long horizons, and (ii) the SVD block, which captures the correlation between multiple channels of the time series. In the context of time-series forecasting, the sequences $\mathbf{X}_{t-T+1: t}$ are used as input into the FT-SVD. The middle layers of the FT and SVD blocks then learn the corresponding parameters in the latent spaces. The output of the network is set to be $\mathbf{X}_{t+1: t+\tau}$, i.e., the predicted sequences. The feature learning of one single FT-SVD layer can be written as:
\begin{equation}
\mathbf{x}_{l+1} =  ~\mathbf{F}^{-1}_l \mathbf{W}_l \mathbf{F}_l \mathbf{x}_l + \sigma ((\mathbf{U}_\mathbf{\Phi} \odot \mathbf{U}_\mathbf{x} )(\mathbf{S}_\mathbf{\Phi} \odot \mathbf{S}_\mathbf{x} )(\mathbf{V}_\mathbf{\Phi} \odot \mathbf{V}_\mathbf{x} ))
\end{equation}
%\begin{equation}
%\begin{aligned}
%\mathbf{x}_{l+1} =  &~\mathbf{F}^{-1}_l \mathbf{W}_l \mathbf{F}_l \mathbf{x}_l \\ &+ \sigma (SVD^{-1}(SVD(\mathbf{\Phi}_l) \odot SVD(\mathbf{x}_l)))
%\end{aligned}
%\end{equation}
%where
%\begin{equation}
%\begin{aligned}
%&SVD^{-1}(SVD(\mathbf{\Phi}_l) \odot SVD(\mathbf{x}_l))\\
%&= (\mathbf{U}_\mathbf{\Phi} \odot \mathbf{U}_\mathbf{x} )(\mathbf{S}_\mathbf{\Phi} \odot \mathbf{S}_\mathbf{x} )(\mathbf{V}_\mathbf{\Phi} \odot \mathbf{V}_\mathbf{x} )
%\end{aligned}
%\end{equation}
Here, $\mathbf{x}_{l}$ is the input of the $l$th layer;  $\mathbf{x}_{l+1}$ is the output of the $l$th layer as well as the input of the $(l+1)$th layer. $\mathbf{F}_l$ and $\mathbf{F}^{-1}_l$ are the forward and inverse FT matrices in the $l$th layer. $\mathbf{W}_l$ and $\mathbf{\Phi}_l$ are the parameters in the Fourier domain and the SVD domain in the $l$th layer. The loss function for the network training is defined as $\mathcal{L}(\mathbf{W}, \boldsymbol{\Phi}) = \frac{1}{BM}\sum_i^B \|\hat{\mathbf{y}}^i(\mathbf{W}, \boldsymbol{\Phi}) - \mathbf{y}^i\|$, where $B$ is the batch size, $M$ is the length of time seires, and $\hat{\mathbf{y}}$ denotes the predicted time series. 

Then we replace the FT blocks and SVD blocks with sparse matrix blocks and Conv. We named them as FT-SVD, FT-Conv, FT-Matrix, Conv-SVD and their corresponding Algorithms (e.g., pseudo-codes) are shown in \textcolor{blue}{S.I. Appendix E}

\section{Experiments}
\label{section: Experiments}

In this section, we test the performance of the four models (FT-Matrix, FT-SVD, FT-Conv, and Conv-SVD) on time-series forecasting based on several real-world datasets. We also compare their performances with those of selected baseline models. The source code is available at \url{https://github.com/Anonymity111222/TLNets}.

\subsection{Datasets}

Experiments were conducted on several real-world benchmark datasets \cite{2020arXiv201207436Z}, which include the Electricity Transformer Temperature (ETT), Electricity, Exchange, Traffic, Weather, and ILI datasets. A comprehensive overview of the datasets can be found in \cite{wu2021autoformer}, and the data source is publically available \footnote{https://github.com/thuml/Autoformer}. It should be noted that ETT comprises four distinct datasets (ETTh1, ETTh2, ETTm1, ETTm2) each containing seven variables. The datasets were split into training, validation, and testing sets using a 7:1:2 ratio. To evaluate our model, we used Mean Absolute Errors (MAE) and Mean Squared Errors (MSE), as done in \cite{2020arXiv201207436Z}. Smaller MAE/MSE values indicate superior model performance. The results presented are the averages of all predictions. Specific details regarding the datasets can be found in Table \ref{tab:datasets}.

\begin{table*}[t!]
\caption{The statistics of all datasets.}
\vspace{-0.15cm}
\begin{center}
% \footnotesize
\scriptsize
\resizebox{\textwidth}{!}
{
\begin{tabular}{c|ccccccc} 
\toprule
Datasets    & ETTh1$\&$ETTh2 & ETTm1 $\&$ETTm2 & Traffic & Electricity & Exchange-Rate & Weather & ILI    \\ 
\midrule
Variates    & 7              & 7               & 862     & 321         & 8             & 21      & 7      \\
Timesteps   & 17,420         & 69,680          & 17,544  & 26,304      & 7,588         & 52,696  & 966    \\
Granularity & 1hour          & 5min            & 1hour   & 1hour       & 1day          & 10min   & 1week  \\
\bottomrule
\end{tabular}}
\end{center}
\label{tab:datasets}
\vspace{-0.2cm}
\end{table*}

\subsection{Baseline Models}
We compare our four TLNets with several baseline models, namely, Informer~\cite{2020arXiv201207436Z}, LogTrans~\cite{li2019enhancing}, Pyraformer*~\cite{ShizhanLiu2023PYRAFORMERLP}, Autoformer~\cite{wu2021autoformer}, FEDformer~\cite{TianZhou2022FEDformerFE}, and the Linear*, NLinear* and DLinear*~\cite{AilingZeng2022AreTE}.

\subsection{Results}

\begin{table*}[t!]
\caption{Multivariate predictions of ETTh1, ETTh2, ETTm1, ETTm2, Traffic, Electricity, Exchange-Rate, Weather and ILI, by nine models. Note: The comparison with other additional methods can be further found in \textcolor{blue}{S.I. Appendix G}}
\centering
\resizebox{1\textwidth}{!}{
\begin{tabular}{c|c|llllllll|cccccccccc} 
\toprule
\multicolumn{2}{c|}{Methods}       & \multicolumn{2}{c|}{FT-Matrix}                                                        & \multicolumn{2}{c|}{FT-SVD}                                                           & \multicolumn{2}{c|}{FT-Conv}                      & \multicolumn{2}{c|}{Conv-SVD}                     & \multicolumn{2}{c|}{Linear*}                       & \multicolumn{2}{c|}{NLinear*}             & \multicolumn{2}{c|}{DLinear*}             & \multicolumn{2}{c|}{FEDformer}           & \multicolumn{2}{c}{Autoformer}  \\ 
\midrule
\multicolumn{2}{c|}{Metric}        & \multicolumn{1}{c}{MSE}                   & \multicolumn{1}{l|}{MAE}                  & MSE                                       & \multicolumn{1}{l|}{MAE}                  & MSE   & \multicolumn{1}{l|}{MAE}                  & MSE                     & MAE                     & \multicolumn{1}{l}{MSE} & \multicolumn{1}{c|}{MAE} & MSE            & \multicolumn{1}{c|}{MAE} & MSE            & \multicolumn{1}{c|}{MAE} & MSE           & \multicolumn{1}{c|}{MAE} & MSE   & MAE                     \\ 
\midrule
\multirow{4}{*}{Electricity} & 96  & 0.141                                     & 0.234                                     & \textbf{\textbf{0.133}}                   & \textbf{\textbf{0.227}}                   & 0.138 & 0.231                                     & 0.134                   & 0.228                   & \uline{0.140}           & \uline{0.237}            & 0.141          & \uline{0.237}            & \uline{0.140}  & \uline{0.237}            & 0.193         & 0.308                    & 0.201 & 0.317                   \\
\bottomrule
\end{tabular}}
\label{table:data-results}
\end{table*}

Table \ref{table:data-results} summarizes the prediction results for nine datasets. Our model performed exceptionally well on most datasets, with the exception of the ILI dataset (although it did have the second-best performance). The small size of the ILI dataset restricted its ability to effectively measure the efficacy of a model. Nevertheless, our models' success on multiple datasets confirms RFL's practical application. FT-SVD is a global receptive field model, while FT-Matrix, FT-Conv, and Conv-SVD incorporate both local and global receptive fields. However, the amount of training data and input sequence length influenced the models' performance. To optimize the models' performance under both factors, ETTm1, ETTm2, Traffic, Electricity, and Weather were trained with an input length of 1440 when predicting a 720-point horizon. All other datasets were trained using an input length of 336 (ILI was trained with an input length of 104).

\begin{figure*}[t!]
\centering
\includegraphics[width=0.99\textwidth]{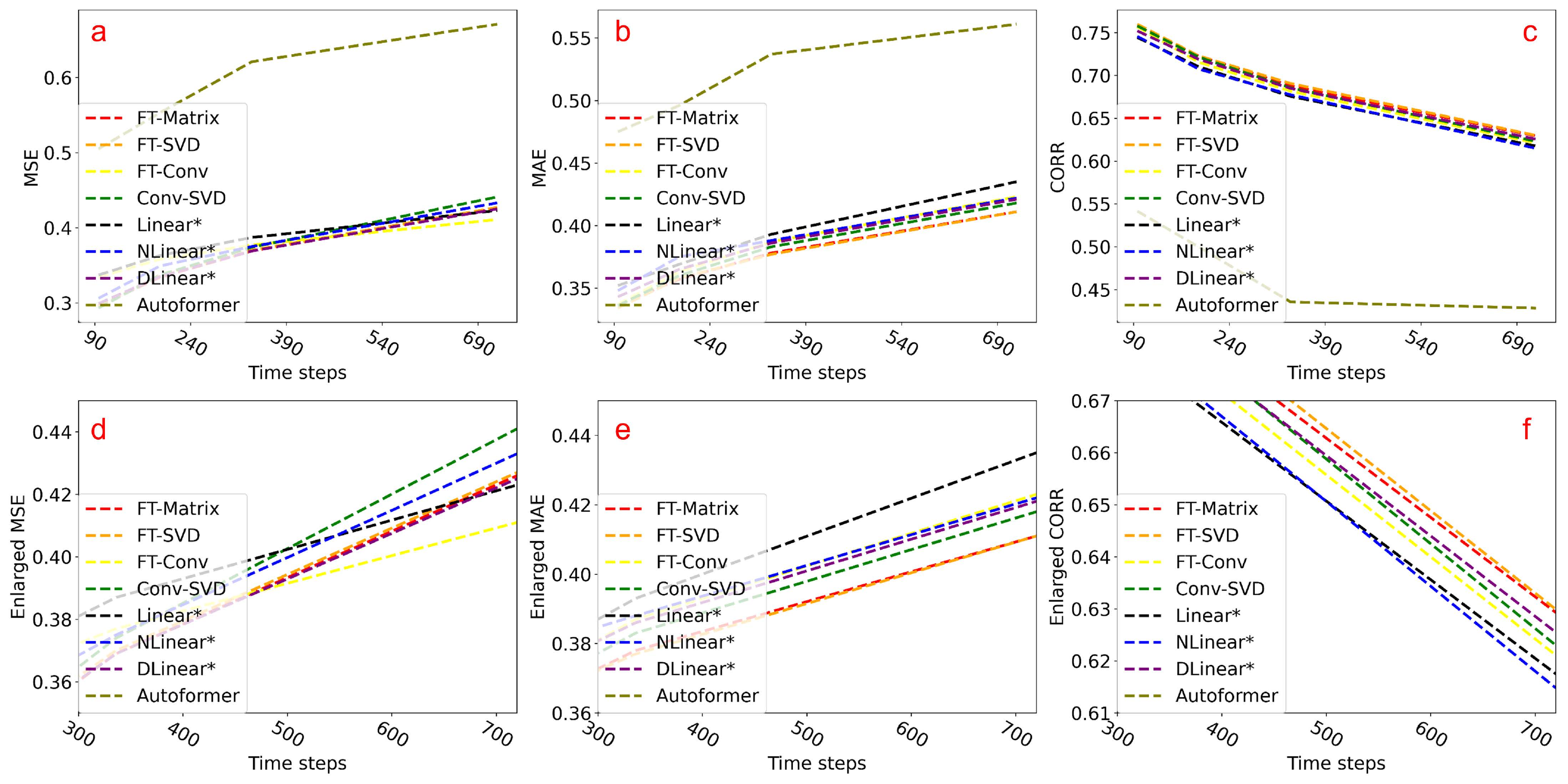}  
\caption{The comparison of MSE, MAE and CORR of ETTm1 with different prediction time steps in a, b and c, while the c, d and e zoom in on the selected portions of the graph.}   
\label{fig:ETTm1_metrics} 
\end{figure*}

Figure \ref{fig:ETTm1_metrics} compares the MSE (Mean Squared Error), MAE (Mean Absolute Error), and CORR (Correlation Coefficient) of ETTm1. Lower MSE and MAE values signify better performance, while higher CORR values indicate a stronger correlation between predicted and actual data. Clearly, our models outperformed others, particularly FT-SVD and FT-Matrix models.

In \textcolor{blue}{S.I. Appendix F}, we provide a comprehensive breakdown of the prediction results and variate 1 distribution of various models, including FT-SVD, FT-Matrix, FT-Conv, Conv-SVD, DLinear, NLinear, and Autoformer models, on the ETTm1 dataset. Overall, our model demonstrated superior performance compared with others.

% \begin{figure*}[t!]
% \centering
% \includegraphics[width=0.9\textwidth]{ETTm1_pred_192_192.pdf}  
% \caption{The prediction results (Horizon = 192; Variate 1) of FT-Matrix, FT-SVD, FT-Conv, Conv-SVD, DLinear*, and NLinear*, Autoformer on the ETTm1 dataset.}   
% \label{fig:ETTm1_pred_192} 
% \end{figure*}

% \begin{figure*}[t!]
% \centering
% \includegraphics[width=0.9\textwidth]{ETTm1_1440_pre_720.pdf}  
% \caption{The prediction results (Horizon = 720; Variate 1) of FT-Matrix, FT-SVD, FT-Conv, Conv-SVD, DLinear*, and NLinear*, Autoformer on the ETTm1 dataset.}   
% \label{fig:ETTm1_pred_720} 
% \end{figure*}

If one wants to use the SVD block on univariate predictions, this can be realized by translating the 1D data into high-dimensional features by Conv1D before prediction. The prediction results of the univariate cases can be found in \textcolor{blue}{S.I. Appendix D}.

\paragraph{Ablation Experiment:}
We herein perform ablation study on TLNets. We trained networks with each building block seperately, namely FT, matrix, SVD, and Conv. The results are shown in Table \ref{table:ablation}. It is evident that the combination of these blocks in one network yields a better performance.

\begin{table}[t!]
\caption{Ablation study of TLNets on multivariate predictions of ETTh1. Here, TLNets best represents the best model among FT-SVD, FT-Matrix, FT-Conv, and Conv-SVD.}
\vspace{6pt}
\centering
\begin{footnotesize}
%\resizebox{1\textwidth}{!}{
\begin{tabular}{c|llcccccccc} 
\toprule
Methods & \multicolumn{2}{c|}{TLNets best}                                                      & \multicolumn{2}{c|}{Matrix}                        & \multicolumn{2}{c|}{FT}          & \multicolumn{2}{c|}{SVD}         & \multicolumn{2}{c}{Conv}  \\ 
\midrule
Metric  & \multicolumn{1}{c}{MSE}                   & \multicolumn{1}{l|}{MAE}                  & \multicolumn{1}{l}{MSE} & \multicolumn{1}{c|}{MAE} & MSE   & \multicolumn{1}{c|}{MAE} & MSE   & \multicolumn{1}{c|}{MAE} & MSE   & MAE               \\ 
\midrule
96      & \textbf{\textbf{0.366}}                   & \textbf{\textbf{0.388}}                   & 0.686                   & 0.544                    & 0.374 & 0.397                    & 0.892 & 0.584                    & 0.388 & 0.418             \\
192     & \textbf{0.408}                            & \textbf{\textbf{0.414}}                   & 0.701                   & 0.559                    & 0.454 & 0.456                    & 0.917 & 0.607                    & 0.432 & 0.448             \\
336     & \textbf{0.433}                            & \textbf{0.423}                            & 0.705                   & 0.570                    & 0.465 & 0.450                    & 0.903 & 0.618                    & 0.460 & 0.465             \\
720     & \textbf{\textbf{\textbf{\textbf{0.423}}}} & \textbf{\textbf{\textbf{\textbf{0.440}}}} & 0.719                   & 0.596                    & 0.459 & 0.462                    & 0.923 & 0.641                    & 0.507 & 0.510             \\
\bottomrule
\end{tabular}%}
\end{footnotesize}
\label{table:ablation}
\vspace{-6pt}
\end{table}

\vspace{-3pt}
\subsection{Complexity Analysis}
We herein provide the complexity analysis of the proposed TLNets. The complexity of SVD is generally $O(nk^2 + k^3 )$, where $n$ is the dimension in time series and $k$ is the number of features. The complexity of FFT and IFFT is $O(n\log n)$. So the complexity of FT, for all features, is $O(nk\log n)$. The total complexity for the summation of Fourier and SVD blocks is $O(nk^2 + k^3 + nk\log n)$. The complexity of matrix multiplication is $O(onk)$, where $o$ is the number of output dimensions.

\vspace{-3pt}
\section{Conclusion}
\label{section: Conclusion}

In this paper, we investigate the problem of time-series forecasting. Specifically, we give a definition to RFL and leverage our prior knowledge of signal processing and deduction about convolution to design four new learning models for time-series forecasting called FT-Matrix, FT-SVD, FT-Conv, and Conv-SVD. These models consist of Fourier, SVD, matrix multiplication, and Conv transformations as basic building blocks for the networks. The Fourier block learns the dominant frequency contents that reflect time-series variation at both short and long horizons. The SVD block captures the correlation between multiple channels of the time series and learns multivariate dynamics in orthogonal spaces. Sparse matrix multiplication can learn local and global information depending on the specific design that enalbes flexibility. The convolution operation can learn the local information. 

We also present an explanation from a new perspective of the link between FT and convolution. We prove that to meet the requirement of a larger receptive field by traditional Conv-based neural networks with small kernels, the network must be deep and take advantage of the activation function to disrupt the pattern of convolutional kernels in the same layers. The proper balance between local and global receptive field for Conv-based networks essentially becomes a bottleneck problem. To achieve a larger receptive field learning capacity, we start from the basics of FT and matrix multiplication and derive their connection with convolution. Then, we design several models of TLNets that project feature learning into interpretable latent spaces through specified transformations. The proposed models have been extensively tested and compared with multiple baseline models using several real-world datasets. Results demonstrate that TLNets outperform existing state-of-the-art methods and show great potential for long-range time-series forecasting.

\vspace{-3pt}
\section{Limitations}

Although we have provided an explanation of convolution and introduced the concept of RFL in deep learning, there is still much work to be done. Firstly, while the deep learning process can be viewed as RFL, it is challenging to design an effective and efficient transformation that works for general tasks. Currently, we typically rely on transformations based on our prior knowledge. For example, incorporating multiple transformations or generating the corresponding matrix based on the data may improve performance. Secondly, the computation requirements of FT-Matrix are influenced by both the input length and the number of features, which means that an increase in the number of features may require more computational resources. This issue could be addressed by designing a more effective sparse matrix or by replacing the transformation with another function with similar properties. We will address these issues in our future study.

\clearpage
\bibliographystyle{unsrt}
\begin{small}
\bibliography{refs}
\end{small}

%\clearpage

% \section{Introduction}\label{s:introduction}

% \section{Conclusion}\label{s:conclusion}

%\section*{Acknowledgments}
%The work is supported by the xxxx project (to be added). The datasets and codes uased in this study are available at https://xxxxx (to be added).

%\section*{Data Availability}
%All the datasets and source codes to reproduce the results in this study are available on GitHub at \url{xxxxx}.

%\section*{References}
% \bibliographystyle{unsrt}
% \begin{small}
% \bibliography{refs}
% \end{small}

\newpage 
\appendix 

\section*{APPENDIX}

\section{Relationship between Fourier Transformation and Matrix Multiplication}\label{section: The Relationship between Fourier Transformation and Matrix Multiplication}
In Section \ref{section: 1D DFT and Convolution}, we provide the general definition of 1D Discrete Fourier Transform (DFT) and Convolution. We then proceed to prove the convolution theory using circular convolution in Section \ref{section: Circular Convolution and Convolution Theory}. Unlike the commonly used convolution in CNN, circular convolution satisfies the convolution theory, making it easier for us to study certain problems. Furthermore, circular convolution closely approximates the common convolution used in CNN, with only the first two elements being different. 
%In Section \ref{Section: The Changes of Frequencies}, we utilize the convolution theory to demonstrate the changes of frequencies in a network.

\subsection{1D Discrete FT (DFT) and Convolution}\label{section: 1D DFT and Convolution}

In practical applications and experiments, the most frequently used method for Fourier Transform is the Discrete Fourier Transform (DFT), owing to the computer's limitations in recording continuous data. Therefore, our goal is to illustrate the connection between the discrete versions of Fourier Transform and convolution.

%\textbf{\textit{Definition 1} -- 1D DFT:}
\begin{equation}
\begin{aligned}
\operatorname{DFT}[x(n)](k)=X(k)=\sum_{n=0}^{N-1} x(n) e^{-i 2 \pi k n / N}
\end{aligned}
\end{equation}
where $N$ is the number of samples in the original domain; 
$n$ the number of current samples;
$k$ the current frequency $k \in[0, N-1]$;
$x_{n}$ the discrete sequence;
and $X_{k}$ the DFT of $x_{n}$ when the frequency equal to $k$.

%\textbf{\textit{Definition 2} -- 1D Inverse DFT:}
\begin{equation}
I D F T[X(k)](n)=x(n)=\frac{1}{N} \sum_{k=0}^{K} X(k) e^{i 2 \pi k n / N}
\end{equation}
where $K$ is the number of samples in the Fourier domain (frequency domain).

%\textbf{\textit{Definition 3} -- Discrete Convolution:}
\begin{equation}
\begin{aligned}
H(n)=[\mathbf{x} * \mathbf{h}](n)=\sum_{j=0}^{M-1} x(n-j) h(j)
\end{aligned}
\end{equation}
where $\mathbf{x}$ and $\mathbf{h}$ are two sequences; $*$ is the convolution operation.

\subsection{Circular Convolution and Convolution Theory}\label{section: Circular Convolution and Convolution Theory}

In this section, we primarily reference the deduction found on the website\footnote{https://zhuanlan.zhihu.com/p/176935055}. Assuming that the size of vectors $\mathbf{x}$ and $\mathbf{h}$ are equal. So the circular convolution can be written as:

\begin{equation}
\begin{aligned}
H(n)=[\mathbf{x} \circledast \mathbf{h}](n)=\sum_{m=0}^{M-1} x(m) h[(n-m)mod N]
\end{aligned}
\end{equation}

where $\circledast$ is circular convolution. We write the circular convolution in matrix format.

\begin{equation}
\begin{aligned}
\mathbf{x} \circledast \mathbf{h}=\left[\begin{array}{lllll}
x_0 & x_{N-1} & x_{N-2} & \cdots & x_{1}\\
x_1 & x_0 & x_{N-1} & \cdots & x_{2}\\
\vdots  & \vdots & \vdots  & \cdots & \vdots \\
x_{N-1} & x_{N-2} & x_{N-3} & \cdots & x_{0}
\end{array}\right]\left[\begin{array}{l}
z_0 \\
z_1 \\
\vdots \\
z_{N-1}
\end{array}\right]
\end{aligned}
\label{Eq: circle_conv}
\end{equation}

Considering the Eq. \eqref{Eq: circle_conv}, we give the following matrix:

\begin{equation}
\begin{aligned}
C(\mathbf{x})=\left[\begin{array}{ccccc}
x_0 & x_1 & x_2 & \cdots & x_{N-1} \\
x_{N-1} & x_0 & x_1 & \cdots & x_{N-2} \\
x_{N-2} & x_{N-1} & x_0 & \cdots & x_{N-3} \\
\vdots & \vdots & \ddots & \ddots & \vdots \\
x_1 & x_2 & \cdots & x_{N-1} & x_0
\end{array}\right]
\end{aligned}
\label{Eq: C-x}
\end{equation}

So the equation Eq. \eqref{Eq: circle_conv} can be rewritten as $x \circledast h = C(x)^T h$. 

We define the permutation matrix as:

\begin{equation}
\mathbf{P}=\left[\begin{array}{ccccc}
0 & 0 & 0 & \cdots & 1 \\
1 & 0 & 0 & \cdots & 0 \\
0 & 1 & 0 & \cdots & 0 \\
\vdots & \vdots & \ddots & \ddots & \vdots \\
0 & 0 & \cdots & 1 & 0
\end{array}\right]
\label{Eq: p}
\end{equation}

So Fourier transformation matrix $\mathbf{F}$ is the combination of the eigenvalue of $\mathbf{P}$. The eigenvalues are $[1, w^{-1}, w^{-2}, \cdots, w^{-(N-1)}]$, where $w=e^{-i 2 \pi / N}$. We have

\begin{equation}
\begin{aligned}
\mathbf{P} \mathbf{F} &=\mathbf{F} \operatorname{diag}\left(\left[1, w^{(N-1) \cdot 1}, w^{(N-1) \cdot 2}, \cdots, w^{(N-1) \cdot(N-1)}\right]^T\right) \\
&=\mathbf{F} \operatorname{diag}\left(\left[1, w^{-1}, w^{-2}, \cdots, w^{-(N-1)}\right]^T\right)
\end{aligned}
\label{Eq: PF}
\end{equation}

$\mathbf{P}$ to the power of $n$.

\begin{equation}
\mathbf{P}^n=\mathbf{F} {d i a g}^n\left(\left[1, w^{-1}, w^{-2}, \cdots, w^{-(N-1)}\right]^T\right) \mathbf{F}^{-1}
\label{Eq: P^n}
\end{equation}

Based on Eq. \eqref{Eq: C-x}, \eqref{Eq: p}, \eqref{Eq: PF}, and \eqref{Eq: P^n}, we could have the following deduction:

\begin{equation}
\begin{aligned}
&C^T(\mathbf{x})=x_0 \mathbf{I}+x_1 \mathbf{P}+x_2 \mathbf{P}^2+\cdots+x_{N-1} \mathbf{P}^{N-1} \\
&=x_0 \mathbf{I}+x_1 \mathbf{F} \operatorname{diag}\left(\left[1, w^{-1}, w^{-2}, \cdots, w^{-(N-1)}\right]^T\right) \mathbf{F}^{-1}\\
&+x_2 \operatorname{\mathbf{F}diag}^2\left(\left[1, w^{-1}, w^{-2}, \cdots, w^{-(N-1)}\right]^T\right) \mathbf{F}^{-1} \\
&+x_{N-1} \mathbf{F} \text { diag }^{N-1}\left(\left[1, w^{-1}, w^{-2}, \cdots, w^{-(N-1)}\right]^T\right)\mathbf{F}^{-1}  \\
&=\mathbf{F} \cdot\left(\sum_{n=1}^N x_n \operatorname{diag}^{n-1}\left(\left[1, w^{-1}, w^{-2}, \cdots, w^{-(N-1)}\right]^T\right)\right) \cdot \mathbf{F}^{-1} \\
&=\mathbf{F} diag[\sum_{n=1}^N x_n, \sum_{n=1}^N x_n w^{-(n-1)}, \sum_{n=1}^N x_n w^{-2(n-1)}, \cdots, \sum_{n=1}^N x_n w^{-(N-1)(N-1)}]^T \cdot \mathbf{F}^{-1} \\
&
\end{aligned}
\end{equation}
Let $\mathbf{x}'=[\sum_{n=1}^N x_n, \sum_{n=1}^N x_n w^{-(n-1)}, \sum_{n=1}^N x_n w^{-2(n-1)}, \cdots, \sum_{n=1}^N x_n w^{-(N-1)(n-1)]^T}$.

\begin{equation}
\begin{aligned}
\mathbf{x}'=\left[\begin{array}{ccccc}
1 & 1 & 1 & \cdots & 1 \\
1 & w^{-1 \cdot 1} & w^{-1 \cdot 2} & \cdots & w^{-1 \cdot(N-1)} \\
1 & w^{-2 \cdot 1} & w^{-2 \cdot 2} & \cdots & w^{-2 \cdot(N-1)} \\
\vdots & \vdots & \ddots & \ddots & \vdots \\
1 & w^{-(N-1) \cdot 1} & \cdots & w^{-(N-1) \cdot(N-2)} & w^{-(N-1) \cdot(N-1)}
\end{array}\right]\left[\begin{array}{c}
x_1 \\
x_2 \\
x_3 \\
\vdots \\
x_k
\end{array}\right]=\mathbf{F}^* \mathbf{x}=\hat{\mathbf{x}}^*
\end{aligned}
\end{equation}

We could conclude that $C^T(\mathbf{x})=\mathbf{F} \operatorname{diag}\left(\hat{\mathbf{x}}^*\right) \mathbf{F}^{-1}$. Next, we will give the deduction about circular convolution and Fourier transformation.

\begin{equation}
\begin{aligned}
&C^T(\mathbf{x}) \mathbf{h}=\mathbf{F} \operatorname{diag}\left(\hat{\mathbf{x}}^*\right) \mathbf{F}^{-1} \mathbf{h} \\
&=\mathbf{F} \operatorname{diag}\left(\hat{\mathbf{x}}^*\right) \frac{1}{N} \mathbf{F}^* \mathbf{h} \\
&=\mathbf{F} \operatorname{diag}\left(\hat{\mathbf{x}}^*\right) \frac{1}{N} \hat{\mathbf{h}}^* \\
&
\end{aligned}
\end{equation}

\begin{equation}
\begin{aligned}
\mathbf{F}^*[C^T(\mathbf{x}) \mathbf{h}]&=\mathbf{F}^*[\mathbf{F} \operatorname{diag}\left(\hat{\mathbf{x}}^*\right) \mathbf{F}^{-1} \mathbf{h}] \\
&= \operatorname{diag}\left(\hat{\mathbf{x}}^*\right) \mathbf{h}^* \\
&=\hat{\mathbf{x}}^* \odot \hat{\mathbf{h}}^* \\
&
\end{aligned}
\end{equation}

Taking the conjugate transformation,  We can get the circular convolution theorem:

\begin{equation}
\begin{aligned}
\mathbf{F} C^T(\mathbf{x}) \mathbf{h} =
\mathbf{F}[\mathbf{x}\circledast \mathbf{h}]=\hat{\mathbf{x}} \odot \hat{\mathbf{h}}
\end{aligned}
\end{equation}

\section{The Relationship between Convolution and Matrix Multiplication}
\label{sup_section: The Relationship between Convolution and Matrix Multiplication}

\begin{figure}[t!]
\centering
\includegraphics[width=0.9\textwidth]{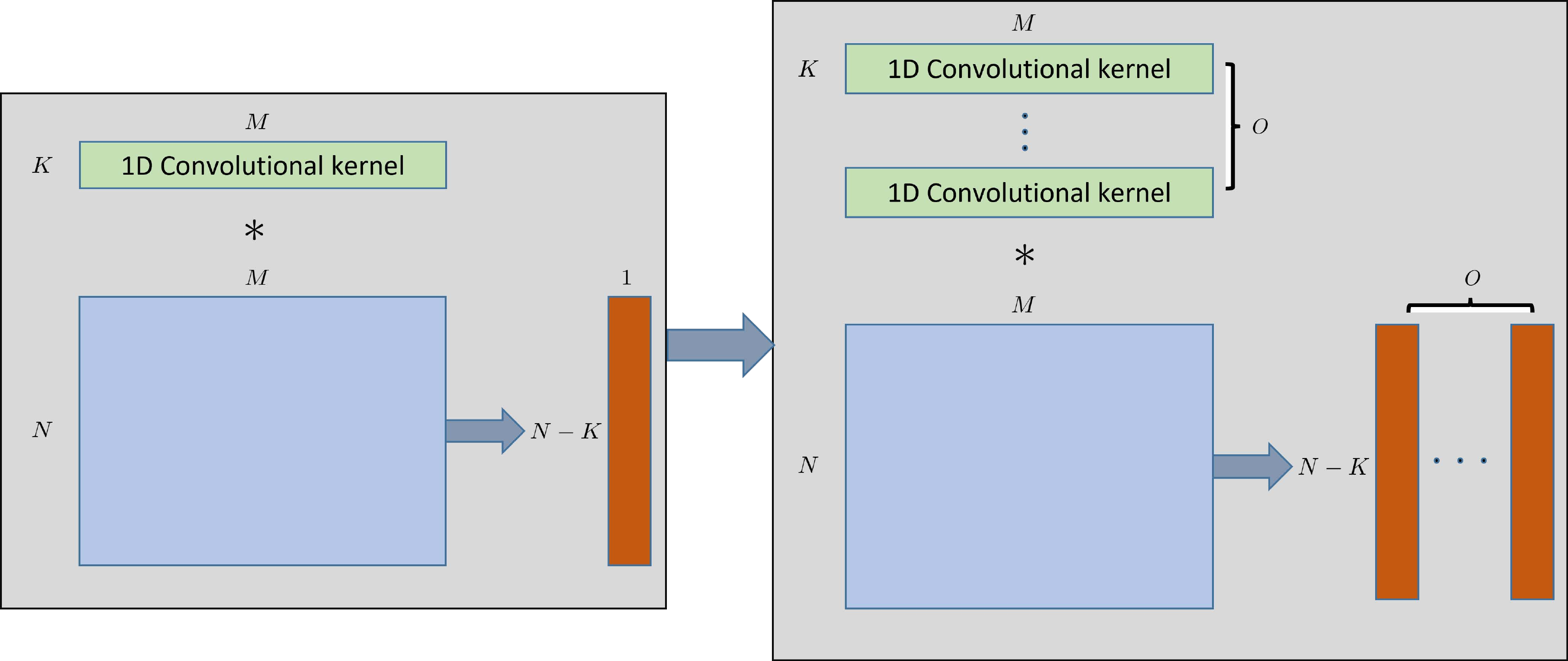}
\caption{The process of 1D convolution. $N$ is the length of the series, $M$ is the number of features, $K$ is the number of kernels and $O$ is the number of output features.}    
\label{fig:Conv1D}
\end{figure}

\begin{figure}[t!]
\centering
\includegraphics[width=0.9\textwidth]{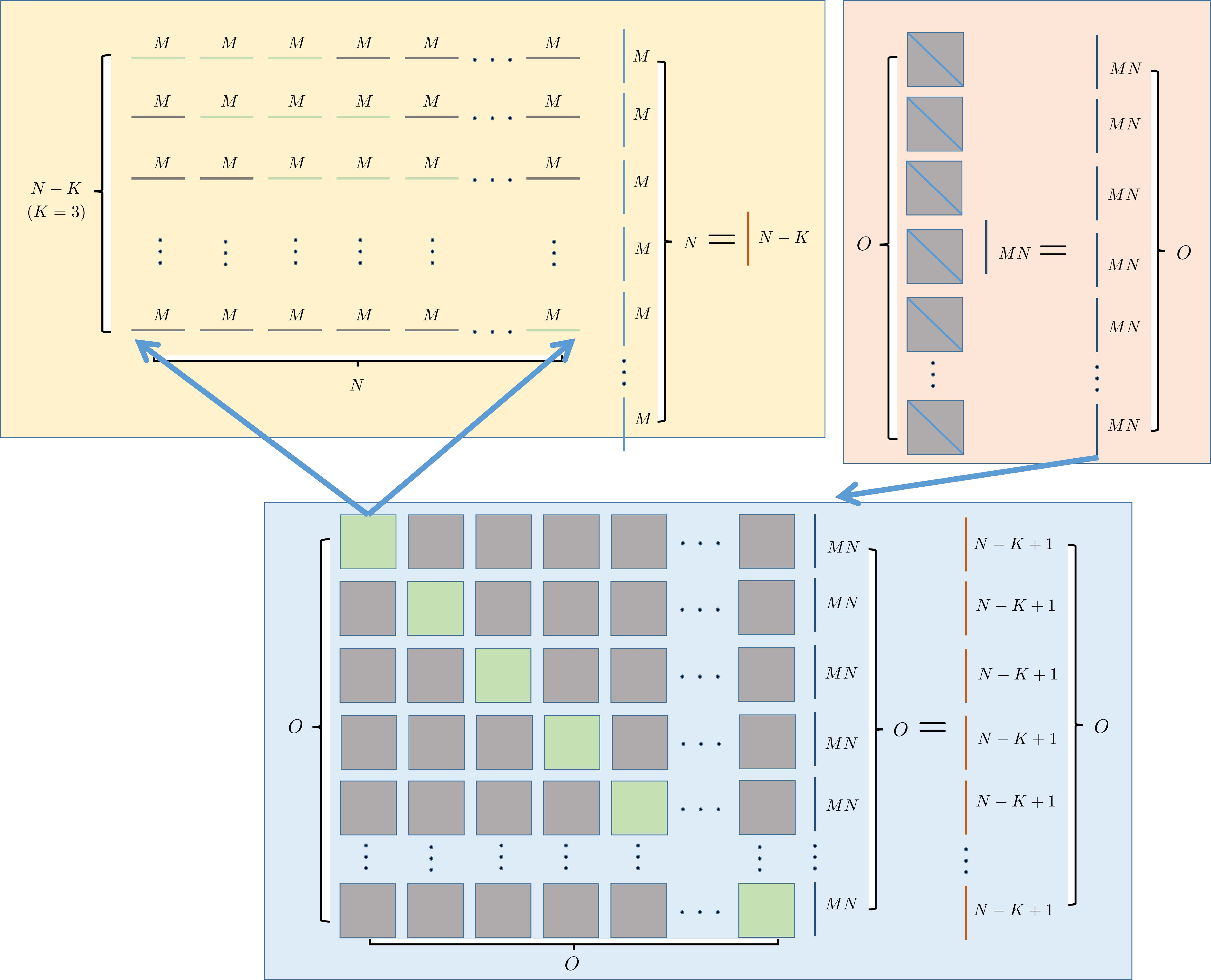}
\caption{The representation of the process of 1D convolution is shown in Figure \ref{fig:Conv1D}. In the top left corner of the figure, we see a basic 1D convolution with a single kernel. The blue lines represent the flattened input, while the green and grey lines represent vectors $\mathbf{1}$ and $\mathbf{0}$ respectively. This left portion of the figure corresponds to the left process in Figure \ref{fig:Conv1D}, and we refer to the left matrix as the kernel matrix.
In the top right corner of the figure, we observe the input being replicated $O$ times, as sometimes it is necessary to enlarge the output size. In the center of the figure, the middle blue box shows the process of convolution with $O$ output channels. The green boxes depict kernel matrices, while the grey boxes represent zero matrices. In this process, the results are convolved by $O$ kernels.}    
\label{fig:Conv1D_all}
\end{figure}

Figure \ref{fig:Conv1D} illustrates the general process of 1D convolution, while Figure \ref{fig:Conv1D_all} presents 1D convolution expressed as matrix multiplication. Firstly, we flatten the data into $(MN, 1)$. Then, we copy it using the unit matrix $O$ times in the first dimension to satisfy the output shape requirement. Finally, we use block matrices to multiply the copied matrix, where the green blocks represent convolutional matrices decided by kernel shape and grey boxes are zero matrices.

By expressing Figure \ref{fig:Conv1D} in the form of Figure \ref{fig:Conv1D_all}, we can describe the process of a 1D convolutional network as:

\begin{equation}
\begin{aligned}
\mathbf{y} =
\sigma\{\mathbf{h}_n...\sigma[\mathbf{h}_1\sigma(\mathbf{h}_1\mathbf{X})]\}
\end{aligned}
\end{equation}

where $\mathbf{h}_1...\mathbf{h}_n$ donate the convolutional matrices in the network. $\mathbf{X}$ denotes the input to the model, while $\mathbf{y}$ refers to the model's output.

Because of the existence of activation functions, the computation must follow a fixed order from front to back which can be seen as ordered matrix multiplication.

\subsection{The Universal Approximation Theory and Mulyi-layer Convolution}

In this section, we will utilize the matrix representation of convolution to elucidate the relationship between universal approximation theory and multi-layer convolution. The universal approximation theory has been proven in \cite{Cybenko_2007}, confirming the convergence of a single-layer neural network. However, the question remains on how to enhance the convergence of a multi-layer neural network. In the following text, we will present our proof, following the symbolic notation from \cite{Cybenko_2007}.

In Theorem 2 from \cite{Cybenko_2007}. Let $\sigma$ be any continuous sigmoidal function. Then finite sums of the form
$$
G(x)=\sum_{j=1}^N \alpha_j \sigma\left(y_j^{\mathrm{T}} x+\theta_j\right)
$$
are dense in $C\left(I_n\right)$. $y_j \in \mathbb{R}^n$ and $\alpha_j, \theta \in \mathbb{R}$ are fixed. Given any $f \in C\left(I_n\right)$ and $\varepsilon>0$, there is a sum, $G(x)$, of the above form, for which
$$
|G(x)-f(x)|<\varepsilon \quad \text { for all } \quad x \in I_n .
$$

This means that if the value of $N$ is sufficiently large, a single-layer neural network can be used to approximate any function.

Regarding deep layers networks, we present the deep formula format of ResNet \cite{He_Zhang_Ren_Sun_2015}, as shown in Equation \eqref{resnet}.

\begin{equation}
\begin{aligned}
&\begin{aligned}
G_1(x) & =y_1 x+\sigma\left(y_1^{\prime} x\right) \\
G_2(x) & =y_2 G_1(x)+\sigma\left[y_2^{\prime} G_1(x)\right] \\
& =y_2\left[y_1 x+\sigma\left(y_1^{\prime} x\right)\right]+\sigma\left\{y_2^{\prime}\left[y_1 x+\sigma\left(y_1^{\prime} x\right)\right]\right\} \\
& =y_2 y_1 x+y_2 \sigma\left(y_1^{\prime} x\right)+\sigma\left[y_2^{\prime} y_1 x+y_2^{\prime} \sigma\left(y_1^{\prime} x\right)\right]
\end{aligned}\\
&\begin{aligned}
G_3(x) & =y_3 G_2(x)+\sigma\left[y_3^{\prime} G_2(x)\right] \\
& =y_3\left\{y_2 y_1 x+y_2 \sigma\left(y_1^{\prime} x\right)+\sigma\left[y_2^{\prime} y_1 x+y_2^{\prime} \sigma\left(y_1^{\prime} x\right)\right]\right. \\
&+  \sigma\left\{y_3^{\prime}\left\{y_2 y_1 x+y_2 \sigma\left(y_1^{\prime} x\right)+\sigma\left[y_2^{\prime} y_1 x+y_2^{\prime} \sigma\left(y_1^{\prime} x\right)\right]\right\}\right\} \\
& =y_3 y_2 y_1 x+y_3 y_2 \sigma\left(y_1^{\prime} x\right)+y_3 \sigma\left[y_2^{\prime} y_1 x+y_2^{\prime} \sigma\left(y_1^{\prime} x\right)\right] \\
&+  \sigma\left\{y_3^{\prime} y_2 y_1 x+y_3^{\prime} y_2 \sigma\left(y_1^{\prime} x\right)+y_3^{\prime} \sigma\left[y_2^{\prime} y_1 x+y_2^{\prime} \sigma\left(y_1^{\prime} x\right)\right]\right\}
\end{aligned}
\end{aligned}
\label{resnet}
\end{equation}

Equation \eqref{resnet} presents a three-layer ResNet architecture, with deeper networks similar to it. Here, $G_i$ represents the output of the $ith$ layer, with $\theta_j$ omitted. The transpose of $y_j$ is omitted for the sake of convenience in notation. Notably, the ResNet structure generates bias as it grows deeper, which is why some networks do not include bias in their experiments yet still achieve high performance. To some extent, the ResNet format satisfies the requirements of universal approximation theory. We can attribute the superior performance of deeper networks over single-layer ones to the latter's tendency to train sparse matrices, which are easier to converge compared to dense matrices. This also explains why ResNet consistently outperforms other models

\section{The Traits, Drawbacks and Improvements of Traditional Convolutional Networks}
\label{section: The Traits, Drawbacks and Improvements of Traditional Convolutional Networks}

\subsection{The receptive fields of single-layer convolution}
We have proved that convolution can be written as matrix multiplication. Firstly, we will use the convolutional matrix to reveal the trend of the changes in the receptive field. We use a simple convolution with a kernel size of three and the input is $(N,1)$ can be expressed as:

\begin{equation}
\begin{aligned}
\mathbf{x} \circledast \mathbf{h} & = 
\left[\begin{array}{lllllll}
h_0 & 0 &  0 & \cdots & h_2 & h_1\\
h_1 & h_0  & 0 & \cdots & 0 & h_2\\
\vdots & \vdots & \vdots & \ddots & \vdots & \vdots \\
0 & 0 & 0 & \cdots & h_1 & h_0\\
\end{array}\right]\left[\begin{array}{l}
x_1 \\
x_2 \\
\vdots \\
x_{N-1} \\
\end{array}\right]\\
&= \mathbf{h}' \mathbf{x} 
%= \mathbf{F}^{-1} [  \mathbf{F} \mathbf{h} \odot \mathbf{F} \mathbf{x}]
\end{aligned}
\label{Eq: circle_conv_matrices}
\end{equation}
We can observe that the small receptive field is shortage of the single-layer convolution. Typically, we adopt the kernel size of 3, so most of the values in the convolutional matrix $\mathbf{h}'$ are zeros. 

\subsection{The receptive fields of two-layer convolution}

As neural networks become deeper, their receptive fields will increase in size, but we do not have general knowledge about that. We could use the changes of the convolutional matrix to reflect this characteristic which is shown in Eq. \ref{receptive_field}.

\begin{equation}
\begin{aligned}
&(\mathbf{x} \circledast \mathbf{h})  \circledast \mathbf{h} \\
=&
\left[\begin{array}{llllllll}
h_0 & 0 & 0 & 0 & \cdots & 0 & h_2 & h_1\\
h_1 & h_0 & 0 & 0 & \cdots & 0 & 0 & h_2\\
\vdots & \vdots & \vdots & \vdots & \ddots & \vdots & \vdots & \vdots \\
0 & 0 & 0 & 0 & \cdots & h_2 & h_1 & h_0\\
\end{array}\right]
\left[\begin{array}{llllllll}
h_0 & 0 & 0 & 0 & \cdots & 0 & h_2 & h_1\\
h_1 & h_0 & 0 & 0 & \cdots & 0 & 0 & h_2\\
\vdots & \vdots & \vdots & \vdots & \ddots & \vdots & \vdots & \vdots \\
0 & 0 & 0 & 0 & \cdots & h_2 & h_1 & h_0\\
\end{array}\right]
\left[\begin{array}{l}
x_1 \\
x_2 \\
x_3 \\
x_4 \\
\vdots \\
x_{N-3} \\
x_{N-2} \\
x_{N-1} \\
\end{array}\right]\\
%&= \mathbf{h}' \mathbf{x} 
&= 
\left[\begin{array}{llllllll}
h_0^2 & 0 & 0 & 0 & \cdots & 2 h_1 h_2  & h_0 h_2 + h_1^2 & 2 h_0 h_1\\
2 h_0 h_1 & h_0^2 & 0 & 0 & \cdots & h_2^2 & 2 h_1 h_2 & h_0 h_2 + h_1^2\\
\vdots & \vdots & \vdots & \vdots & \ddots & \vdots & \vdots & \vdots \\
0 & 0 & 0 & 0 & \cdots & 2h_0 h_2 +h_1^2   & 2 h_0 h_1 & h_0 ^2\\
\end{array}\right]
\left[\begin{array}{l}
x_0 \\
x_2 \\
x_3 \\
x_4 \\
\vdots \\
x_{N-3} \\
x_{N-2} \\
x_{N-1} \\
\end{array}\right]
%= \mathbf{F}^{-1} [  \mathbf{F} \mathbf{h} \odot \mathbf{F} \mathbf{x}]
\end{aligned}
\label{receptive_field}
\end{equation}

To simplify the writing process, we will use the same kernel, as it does not affect the receptive field. Additionally, for the sake of convenience, we did not include the activation function in our computations. Because we aim to identify the general trend of receptive field variations through continuous convolution. It is obvious that the receptive field becomes larger and has a regular pattern.

\subsection{The frequency and convolution}

In section \ref{section: The Relationship between Fourier Transformation and Matrix Multiplication}, we talk about the relationship between Fourier transformation and matrix multiplication. Based on those, we reveal the relationship between kernel size and frequencies in the Fourier domain. A convolution with a kernel size of three is shown in equation \eqref{Eq: single_layer}. The matrix $\mathbf{H}$ only contains low-frequency information, as most of the values in $\mathbf{h}$ are zero. The maximum frequency that can be learned from convolution is determined by the kernel size we choose. For instance, if we set the kernel size to three, only a fraction of the low-frequency information can be captured during the convolution process.

\begin{equation}
\begin{aligned}
\mathbf{H} = \mathbf{F} \mathbf{h}=\left[\begin{array}{c}
H_0 \\
H_1 \\
H_2 \\
H_3 \\
\vdots \\
H_K
\end{array}\right]
&=\left[\begin{array}{cccc}
1 & 1 & \cdots & 1 \\
1 & w^{1*1} & \cdots & w^{1* N} \\
1 & w^{2*1} & \cdots & w^{2* N} \\
1 & w^{3*1} & \cdots & w^{3* N} \\
\vdots & \vdots & & \\
1 & w^{K*1} & \cdots & w^{K * N}
\end{array}\right]\left[\begin{array}{c}
h_0 \\
h_1 \\
h_2 \\
0 \\
\vdots \\
0
\end{array}\right]\\
\end{aligned}
\label{Eq: single_layer}
\end{equation}

\section{The Relationship between Multi-head Attention and Matrix Multiplication}

\begin{figure*}[t!]
\centering
\includegraphics[width=1\textwidth]{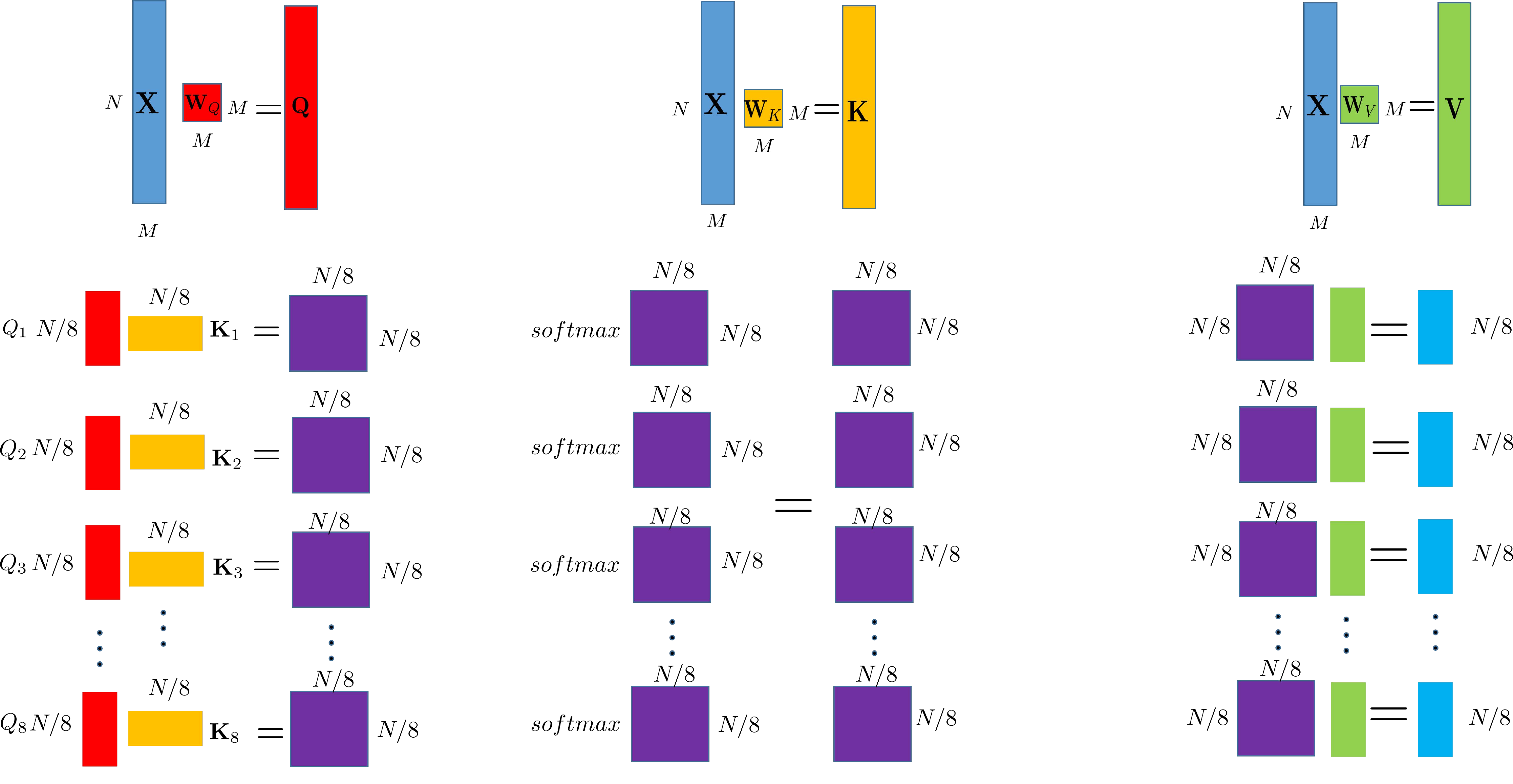}  
\caption{The process of multi-head attention. There are eight heads in this process. We use $\mathbf{QK}$ to multiply a sparse matrix. The grey blocks are zeros and the blue  blocks are a matrix full of ones.}   
\label{fig: Attention} 
\end{figure*}

\begin{figure*}[t!]
\centering
\includegraphics[width=1\textwidth]{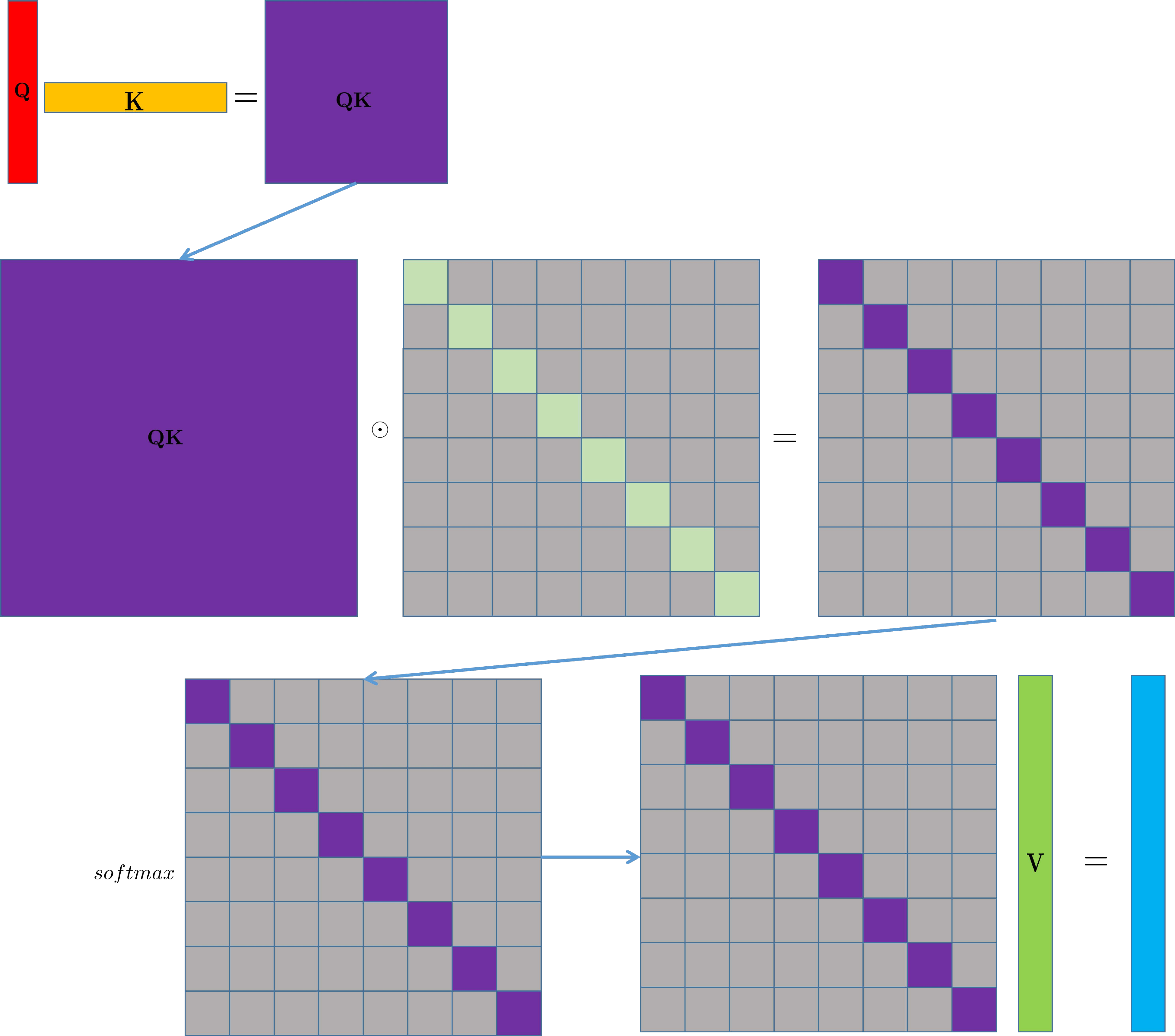}  
\caption{To represent multi-head attention through matrix multiplication, we first calculate the product of $\mathbf{QK}^T$ in the top line. Next, we take the Hadamard product between $\mathbf{QK}$ and a sparse block matrix in which blue boxes represent ones and grey boxes represent zeros, as shown in the second line. Finally, we perform softmax and multiply $\mathbf{V}$ to obtain the output of multi-head attention.}
\label{fig: Attention_Matrix} 
\end{figure*}

The general process of multi-head attention in the transformer is shown in Figure \ref{fig: Attention}, which can be represented by:

\begin{equation}
\begin{aligned}
Attention &= softmax(\frac{\mathbf{Q}\mathbf{K}^T}{\sqrt{d_k}}\mathbf{V})\\
MultiHead(\mathbf{Q},\mathbf{K},\mathbf{V}) &= concat(head_1, ... head_8)\mathbf{W}^O(d_k = d/8)\\
\mathbf{Q} &= \mathbf{X}\mathbf{W}_Q\\
\mathbf{K} &= \mathbf{X}\mathbf{W}_K\\
\mathbf{V} &= \mathbf{X}\mathbf{W}_V\\
\end{aligned}
\label{Eq: att}
\end{equation}

where $head_i = Attention(\mathbf{Q}[i*d_k:(i+1)*d_k], \mathbf{K}[i*d_k:(i+1)*d_k], \mathbf{V}[i*d_k:(i+1)*d_k])$. We represent this process by matrices multiplication in Figure \ref{fig: Attention_Matrix}. According to it the multi-head attention can be rewritten as:

\begin{equation}
\begin{aligned}
&\{softmax[(\mathbf{X}\mathbf{W}_Q)(\mathbf{X}\mathbf{W}_K)^T\odot\mathbf{M} \}(\mathbf{X}\mathbf{W}_V)\\
=&[softmax(\mathbf{X}\mathbf{W}_Q\mathbf{W}_K^T\mathbf{X}^T)\odot\mathbf{M}](\mathbf{X}\mathbf{W}_V)\\
=&[softmax(\mathbf{X}\mathbf{W}_{QK}\mathbf{X}^T)\odot\mathbf{M}](\mathbf{X}\mathbf{W}_V)\\
\end{aligned}
\label{Eq: atten-matrix}
\end{equation}

from Eq. \eqref{Eq: atten-matrix} we could find that multi-head attention is also an ordered matrix multiplication.

\section{The Algorithms of our Models}
In this section, we give the corresponding algorithms of FT-SVD \ref{algo} FT-Matrix \ref{algo: Matrix}, FT-Conv \ref{algo: Conv}, and  Conv-SVD \ref{algo: SVD-Conv}. In FT-Matrix, we randomly initialise parameters $\mathbf{M}$ and then make it multiple a fixed sparse $\mathbf{M}$, because we want to ignore some irrelevant features. $\mathbf{M}$ is filled with $0$ and $1$.  The sizes of convolutional in FT-Conv and Conv-SVD are three.

\begin{algorithm}
\caption{Pseudo-code of FT-SVD}
\KwIn{Input training data $\mathcal{D} = \cup_{i=1}^{n_b} \{\mathbf{x}^i, \mathbf{y}^i\}$; Batch size $B$; Input length $I$; Prediction length $O$; Learning rate $\alpha$. The number of layers $L$.}
\KwOut{Predicted time series  $\hat{\mathbf{y}}^i(\mathbf{W}, \boldsymbol{\Phi})$ }
\KwResult{Optimal parameters $\mathbf{W} = \{\mathbf{W}_{l},\mathbf{W}_{O}\}$ and $\mathbf{\Phi}$.}
\For{j in range~[1, epoch]:}
{\For{i in range~[1, baches]:}
{Assign: $\mathbf{x}_{1} = \mathbf{x}^i$ \; 

\For{l in range~[1, $L-1$]:}
{$\mathbf{x}_{l}^\mathcal{F} = \mathcal{F}^{-1} [\mathbf{W}_{l}^j \mathcal{F} (\mathbf{x}_l)]$ \;

$\{\mathbf{U}_\mathbf{x}, \mathbf{S}_\mathbf{x}, \mathbf{V}_\mathbf{x}\} \leftarrow \texttt{SVD}(\mathbf{x}_l) $ \;

$\{\mathbf{U}_\mathbf{\Phi}, \mathbf{S}_\mathbf{\Phi}, \mathbf{V}_\mathbf{\Phi}\}  \leftarrow \texttt{SVD}(\mathbf{\Phi}_l^j) $ \;

$\mathbf{U}' = \mathbf{U}_\mathbf{x} \odot  \mathbf{U}_\mathbf{\Phi}$,  
$\mathbf{S}' = \mathbf{S}_\mathbf{x} \odot \mathbf{S}_\mathbf{\Phi}$, 
$\mathbf{V}' = \mathbf{V}_\mathbf{x} \odot \mathbf{V}_\mathbf{\Phi} $ \; 

$\mathbf{x}_{l}^{\texttt{SVD}} = \mathbf{U}'\mathbf{S}'\mathbf{V}'$; \; 

$\mathbf{x}_{l} = \mathbf{x}_{l}^\mathcal{F} + \sigma(\mathbf{x}_{l}^{\texttt{SVD}})$ \;}

Output: $\hat{\mathbf{y}}(\mathbf{W}, \boldsymbol{\Phi})= \mathcal{F}^{-1} [\mathbf{W}_{O} \mathcal{F} (\mathbf{x}_{L-1})]$\;

%$\mathbf{y}_i(\mathbf{W}, \boldsymbol{\Phi}) = \mathbf{x}_{out}$\;

Compute: $\mathcal{L}_j$, $\nabla_\mathbf{W} \mathcal{L}_j$, $\nabla_\mathbf{\Phi} \mathcal{L}_j$\;

Update: $\mathbf{W}_l \leftarrow \mathbf{W}_l - \alpha \nabla_\mathbf{W} \mathcal{L}_j$\;

Update: $\mathbf{\Phi}_l \leftarrow \mathbf{\Phi}_l - \alpha \nabla_\mathbf{\Phi} \mathcal{L}_j$\;
}}
\label{algo}
\end{algorithm}

\begin{algorithm}
\caption{Pseudo-code of FT-Matrix}
\KwIn{Input training data $\mathcal{D} = \cup_{i=1}^{n_b} \{\mathbf{x}^i, \mathbf{y}^i\}$; Batch size $B$; Input length $I$; Prediction length $O$; Learning rate $\alpha$. The number of layers $L$. The fixed sparse matrix $\mathbf{M}$}
\KwOut{Predicted time series  $\hat{\mathbf{y}}^i(\mathbf{W}, \boldsymbol{\Phi})$ }
\KwResult{Optimal parameters $\mathbf{W} = \{\mathbf{W}_{l},\mathbf{W}_{O}\}$ and $\mathbf{\Phi}^M$.}
\For{j in range~[1, epoch]:}
{\For{i in range~[1, baches]:}
{Assign: $\mathbf{x}_{1} = \mathbf{x}^i$ \; 

\For{l in range~[1, $L-1$]:}
{$\mathbf{x}_{l}^\mathcal{F} = \mathcal{F}^{-1} [\mathbf{W}_{l} \mathcal{F} (\mathbf{x}_l)]$ \;

$\mathbf{x}_l^M=(\mathbf{M} \odot \mathbf{\Phi}_l^M) \mathbf{x}_l$

$\mathbf{x}_{l} = \mathbf{x}_{l}^\mathcal{F} + \mathbf{x}_l^M$ \;}

Output: $\hat{\mathbf{y}}^i(\mathbf{W}, \boldsymbol{\Phi})= \mathcal{F}^{-1} [\mathbf{W}_{O} \mathcal{F} (\mathbf{x}_{L-1})]$\;

%$\mathbf{y}_i(\mathbf{W}, \boldsymbol{\Phi}) = \mathbf{x}_{out}$\;

Compute: $\mathcal{L}_j$, $\nabla_\mathbf{W} \mathcal{L}_j$, $\nabla_\mathbf{\Phi} \mathcal{L}_j$\;

Update: $\mathbf{W}_l \leftarrow \mathbf{W}_l - \alpha \nabla_\mathbf{W} \mathcal{L}_j$\;

Update: $\mathbf{\Phi}_l \leftarrow \mathbf{\Phi}_l - \alpha \nabla_\mathbf{\Phi} \mathcal{L}_j$\;
}}
\label{algo: Matrix}
\end{algorithm}

\begin{algorithm}
\caption{Pseudo-code of FT-Conv}
\KwIn{Input training data $\mathcal{D} = \cup_{i=1}^{n_b} \{\mathbf{x}^i, \mathbf{y}^i\}$; Batch size $B$; Input length $I$; Prediction length $O$; Learning rate $\alpha$. The number of layers $L$. }
\KwOut{Predicted time series  $\hat{\mathbf{y}}^i(\mathbf{W}, \mathbf{K})$ }
\KwResult{Optimal parameters $\mathbf{W} = \{\mathbf{W}_{l},\mathbf{W}_{O}\}$ and $\mathbf{K} = \{\mathbf{K}_{l},\mathbf{K}_{O}\}$ (Parameters in $Conv$).}
\For{j in range~[1, epoch]:}
{\For{i in range~[1, baches]:}
{Assign: $\mathbf{x}_{1} = \mathbf{x}^i$ \; 

\For{l in range~[1, $L-1$]:}
{$\mathbf{x}_{l}^\mathcal{F} = \mathcal{F}^{-1} [\mathbf{W}_{l}^j \mathcal{F} (\mathbf{x}_l)]$ \;

$\mathbf{x}_l^C=Conv_l(\mathbf{x}_l)$

$\mathbf{x}_{l} = \mathbf{x}_{l}^\mathcal{F} + \mathbf{x}_l^C$ \;}

Output: $\hat{\mathbf{y}}^i(\mathbf{W}, \boldsymbol{\Phi})= \mathcal{F}^{-1} [\mathbf{W}_{O} \mathcal{F} (\mathbf{x}_{L-1})]$\;

%$\mathbf{y}_i(\mathbf{W}, \boldsymbol{\Phi}) = \mathbf{x}_{out}$\;

Compute: $\mathcal{L}_j$, $\nabla_\mathbf{W} \mathcal{L}_j$, $\nabla_\mathbf{\Phi} \mathcal{L}_j$\;

Update: $\mathbf{W}_l \leftarrow \mathbf{W}_l - \alpha \nabla_\mathbf{W} \mathcal{L}_j$\;

Update: $\mathbf{K}_l \leftarrow \mathbf{K}_l - \alpha \nabla_\mathbf{K} \mathcal{L}_j$\;
}}
\label{algo: Conv}
\end{algorithm}

\begin{algorithm}
\caption{Pseudo-code of Conv-SVD}
\KwIn{Input training data $\mathcal{D} = \cup_{i=1}^{n_b} \{\mathbf{x}^i, \mathbf{y}^i\}$; Batch size $B$; Input length $I$; Prediction length $O$; Learning rate $\alpha$. The number of layers $L$.}
\KwOut{Predicted time series  $\hat{\mathbf{y}}^i(\mathbf{K}, \boldsymbol{\Phi})$ }
\KwResult{Optimal parameters $\mathbf{K} = \{\mathbf{K}_{l},\mathbf{K}_{O}\}$ and $\mathbf{\Phi}$.}
\For{j in range~[1, epoch]:}
{\For{i in range~[1, baches]:}
{Assign: $\mathbf{x}_{1} = \mathbf{x}^i$ ;\; 

\For{l in range~[1, $L-1$]:}
{$\mathbf{x}_{l}^C = Conv_l (\mathbf{x}_l)]$ ;\;

$\{\mathbf{U}_\mathbf{x}, \mathbf{S}_\mathbf{x}, \mathbf{V}_\mathbf{x}\} \leftarrow \texttt{SVD}(\mathbf{x}_l) $; \;

$\{\mathbf{U}_\mathbf{\Phi}, \mathbf{S}_\mathbf{\Phi}, \mathbf{V}_\mathbf{\Phi}\}  \leftarrow \texttt{SVD}(\mathbf{\Phi}_l^j) $; \;

$\mathbf{U}' = \mathbf{U}_\mathbf{x} \odot  \mathbf{U}_\mathbf{\Phi}$,  
$\mathbf{S}' = \mathbf{S}_\mathbf{x} \odot \mathbf{S}_\mathbf{\Phi}$, 
$\mathbf{V}' = \mathbf{V}_\mathbf{x} \odot \mathbf{V}_\mathbf{\Phi} $; \; 

$\mathbf{x}_{l}^{\texttt{SVD}} = \mathbf{U}'\mathbf{S}'\mathbf{V}'$; \; 

$\mathbf{x}_{l} = \mathbf{x}_{l}^C + \sigma(\mathbf{x}_{l}^{\texttt{SVD}})$; \;}

Output: $\hat{\mathbf{y}}^i(\mathbf{K}, \boldsymbol{\Phi})=Conv_O (\mathbf{x}_{L-1})]$\;

%$\mathbf{y}_i(\mathbf{W}, \boldsymbol{\Phi}) = \mathbf{x}_{out}$\;

Compute: $\mathcal{L}_j$, $\nabla_\mathbf{K} \mathcal{L}_j$, $\nabla_\mathbf{\Phi} \mathcal{L}_j$\;

Update: $\mathbf{K}^{j} \leftarrow \mathbf{K}^{j} - \alpha \nabla_\mathbf{K} \mathcal{L}_j$\;

Update: $\mathbf{\Phi}^{j} \leftarrow \mathbf{\Phi}^{j} - \alpha \nabla_\mathbf{\Phi} \mathcal{L}_j$\;
}}
\label{algo: SVD-Conv}
\end{algorithm}

\section{The Prediction of ETTm1}
\begin{figure*}[t!]
\centering
\includegraphics[width=0.99\textwidth]{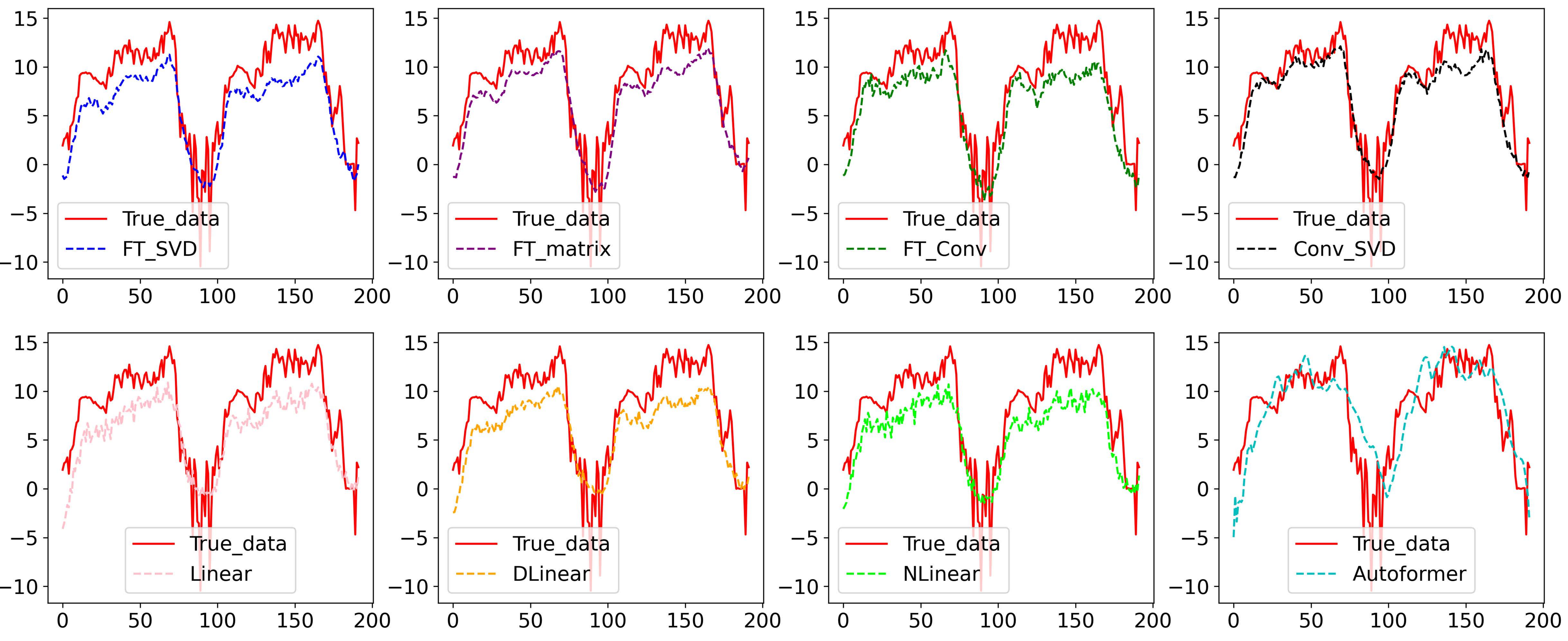}  
\caption{The prediction results (Horizon = 192; Variate 1) of FT-Matrix, FT-SVD, FT-Conv, Conv-SVD, DLinear*, and NLinear*, Autoformer on the ETTm1 dataset.}   
\label{fig:ETTm1_pred_192} 
\end{figure*}

\begin{figure*}[t!]
\centering
\includegraphics[width=0.9\textwidth]{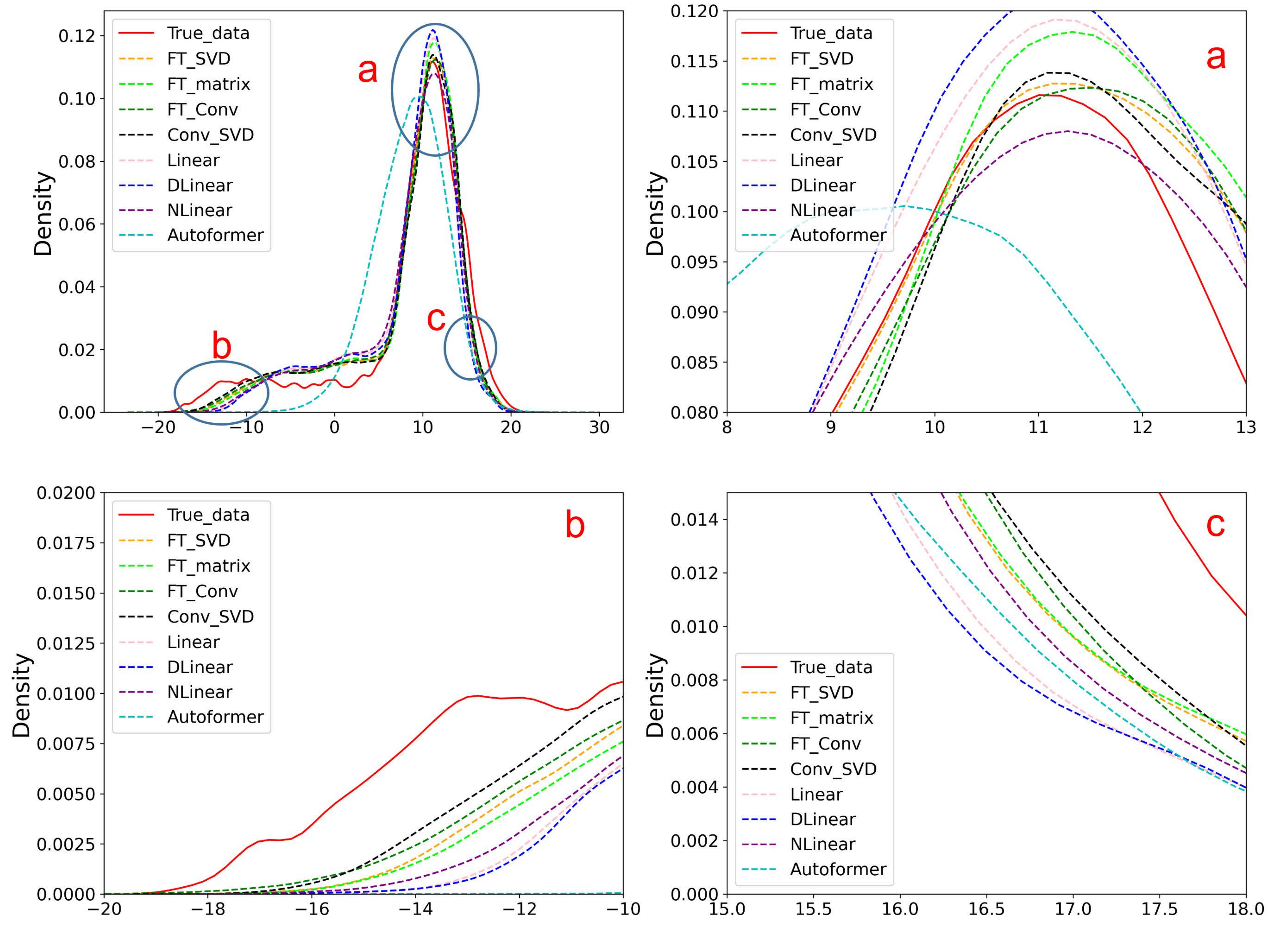}  
\caption{The distribution of Variate 1 of ETTm1 when the prediction horizons are set at 192 and we zoom part of the original image.}   
\label{fig:distrubutions_192} 
\end{figure*}

\begin{figure*}[t!]
\centering
\includegraphics[width=0.99\textwidth]{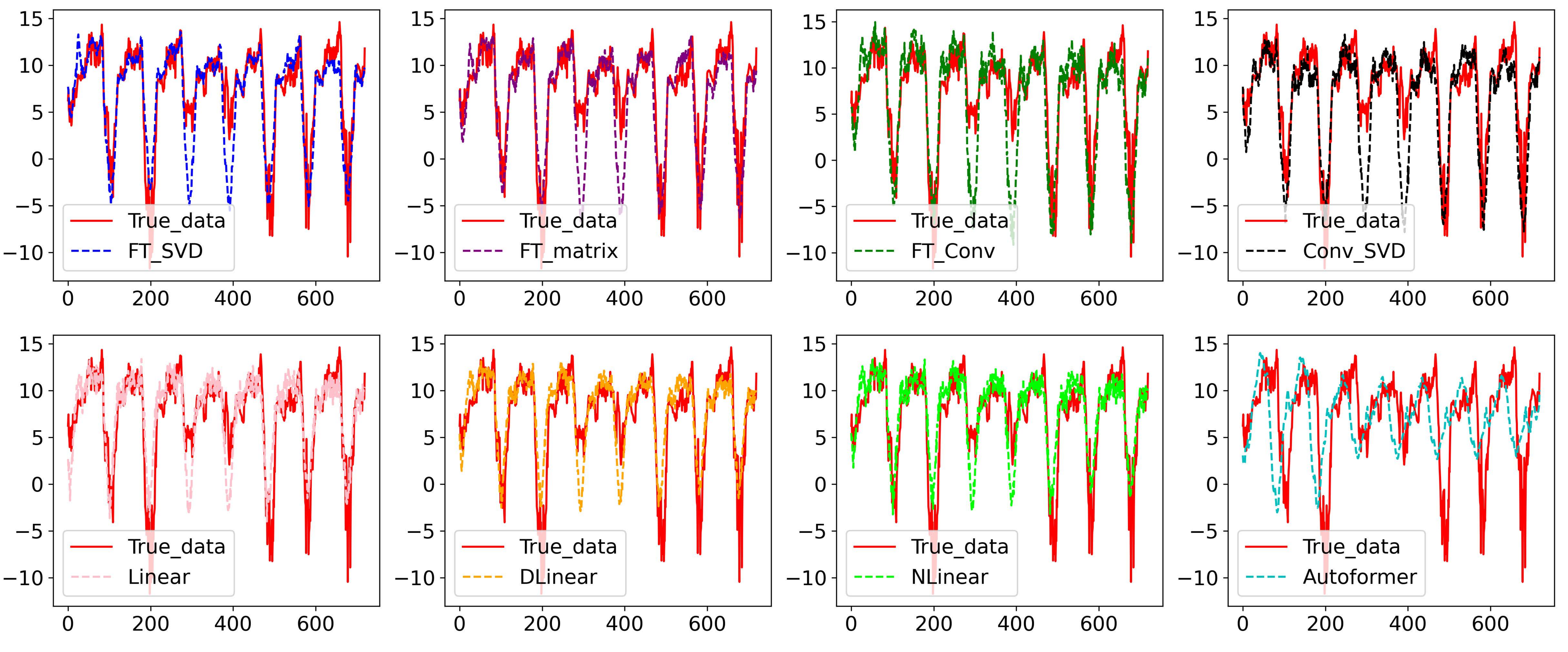}  
\caption{The prediction results (Horizon = 720; Variate 1) of FT-Matrix, FT-SVD, FT-Conv, Conv-SVD, DLinear*, and NLinear*, Autoformer on the ETTm1 dataset.}   
\label{fig:ETTm1_pred_720} 
\end{figure*}

\begin{figure*}[t!]
\centering
\includegraphics[width=0.9\textwidth]{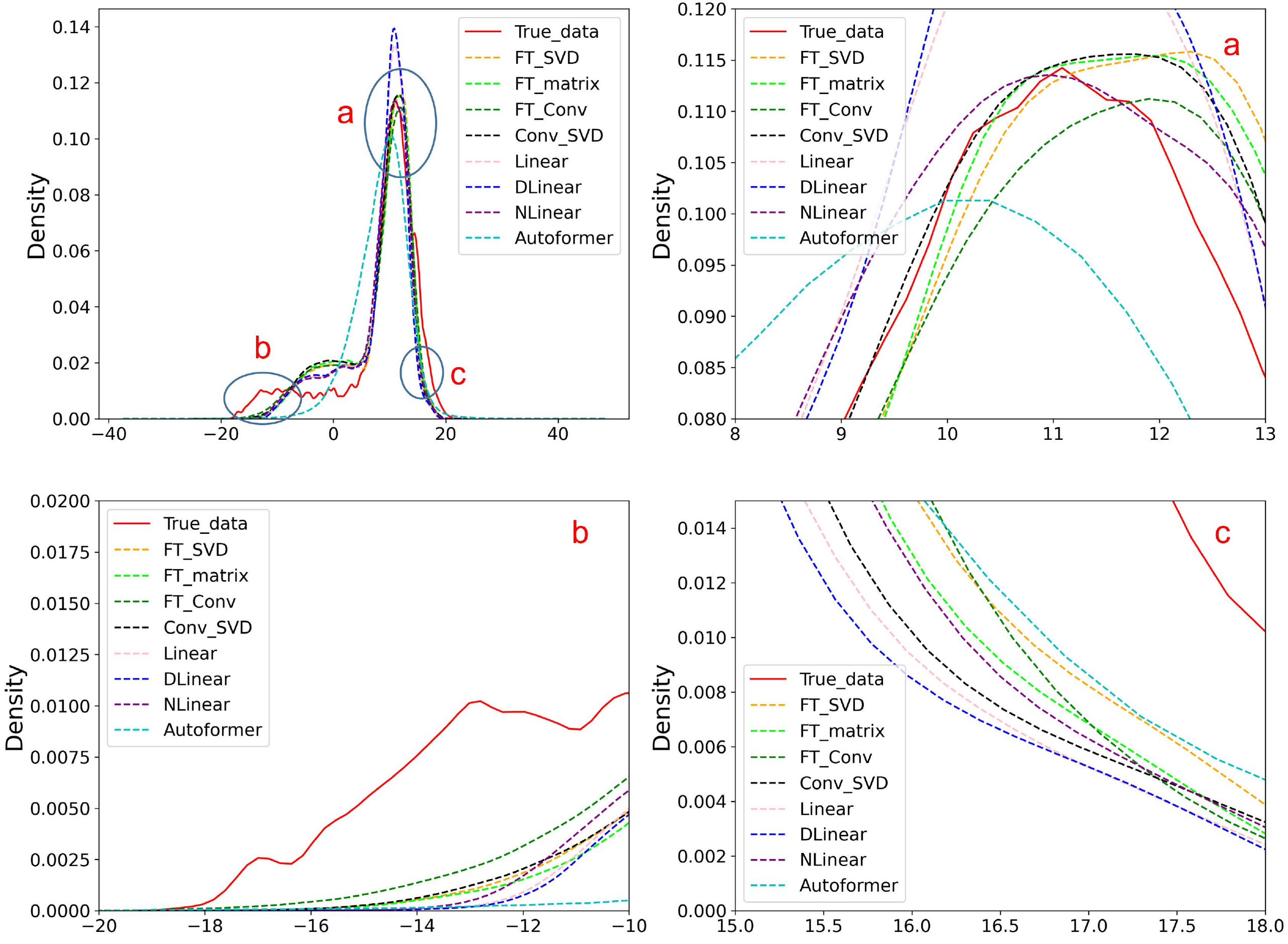}  
\caption{The distribution of Variate 1 of ETTm1 when the prediction horizons are set at 720 and we zoom part of the original image.}   
\label{fig:distrubutions_720} 
\end{figure*}

Figures \ref{fig:ETTm1_pred_192} and \ref{fig:ETTm1_pred_720} show the prediction results of variate 1 of the FT-SVD, FT-Matrix, FT-Conv, Conv-SVD, DLinear, NLinear, and Autoformer models on the ETTm1 dataset. With the exception of Autoformer, all models predicted the trend of the data. In Figure \ref{fig:distrubutions_192} and \ref{fig:distrubutions_720}, we further present the distributions of these models. Our models, FT-SVD and FT-Matrix, more closely align with the true data distribution compared to Linear*, NLinear*, and DLinear*.

\section{Profermances of Prediction}

In this section, We compared the preferences of multivariate and univariate prediction. In general, our results are the best models. In univariate prediction, we expand the one-dimensional feature to four by convolutional firstly when we use FT-SVD and FT-Conv. The results are shown in \ref{sup_table:data-results} and \ref{sup_table:Uni-data-results}.

\begin{table*}[ht]
\caption{Multivariate predictions of ETTh1, ETTh2, ETTm1, ETTm2, Traffic, Electricity, Exchange-Rate, Weather and ILI, by twelve models.}
\centering
\resizebox{0.8\textwidth}{!}{
\rotatebox{90}{
\begin{tabular}{c|c|llllllll|cccccccccccccccc} 
\toprule
\multicolumn{2}{c|}{Methods}       & \multicolumn{2}{c|}{FT-Matrix}                                                        & \multicolumn{2}{c|}{FT-SVD}                                                           & \multicolumn{2}{c|}{FT-Conv}                      & \multicolumn{2}{c|}{Conv-SVD}                     & \multicolumn{2}{c|}{Linear*}             & \multicolumn{2}{c|}{NLinear*}   & \multicolumn{2}{c|}{DLinear*}   & \multicolumn{2}{c|}{FEDformer}           & \multicolumn{2}{c|}{Autoformer} & \multicolumn{2}{c|}{Informer} & \multicolumn{2}{c|}{Pyraformer*} & \multicolumn{2}{c}{LogTrans}  \\ 
\midrule
\multicolumn{2}{c|}{Metric}        & \multicolumn{1}{c|}{MSE}                  & \multicolumn{1}{l|}{MAE}                  & MSE                                       & \multicolumn{1}{l|}{MAE}                  & MSE   & MAE                                       & MSE                     & \multicolumn{1}{l}{MAE} & \multicolumn{1}{c|}{MSE} & MAE           & MSE            & MAE            & MSE            & MAE            & \multicolumn{1}{c|}{MSE} & MAE           & MSE   & MAE                     & MSE   & MAE                   & MSE   & MAE                      & MSE   & MAE                   \\ 
\midrule
\multirow{4}{*}{Electricity} & 96  & 0.141                                     & 0.234                                     & \textbf{\textbf{0.133}}                   & \textbf{\textbf{0.227}}                   & 0.138 & 0.231                                     & 0.134                   & 0.228                   & \uline{0.140}            & \uline{0.237} & 0.141          & \uline{0.237}  & \uline{0.140}  & \uline{0.237}  & 0.193                    & 0.308         & 0.201 & 0.317                   & 0.274 & 0.368                 & 0.386 & 0.449                    & 0.258 & 0.357                 \\
\bottomrule
\end{tabular}}}
\label{sup_table:data-results}
\end{table*}

\begin{table*}[ht]
\caption{Univariate predictions of ETTh1, ETTh2, ETTm1, and ETTm2 by twelve models.}
\centering
\resizebox{0.4 \textwidth}{!}{
\rotatebox{90}{
\begin{tabular}{c|c|llllllll|cccccccccccccccc} 
\toprule
\multicolumn{2}{c|}{Methods}  & \multicolumn{2}{c|}{FT-Matrix} & \multicolumn{2}{c|}{F-SVD}      & \multicolumn{2}{c|}{FT-Conv}    & \multicolumn{2}{c|}{Conv-SVD}            & \multicolumn{2}{c|}{Linear} & \multicolumn{2}{c|}{NLinear}    & \multicolumn{2}{c}{DLinear} & \multicolumn{2}{c|}{FEDformer-f} & \multicolumn{2}{c|}{FEDformer-w} & \multicolumn{2}{c|}{Autoformer} & \multicolumn{2}{c|}{Informer} & \multicolumn{2}{c}{LogTrans}  \\ 
\midrule
\multicolumn{2}{c|}{Metric}   & MSE   & MAE                    & MSE            & MAE            & MSE            & MAE            & MSE                     & MAE            & MSE   & MAE                 & MSE            & MSE            & MAE   & MAE                 & \multicolumn{1}{c|}{MSE} & MAE   & MSE            & MAE             & MSE           & MAE             & MSE   & MAE                   & MSE   & MAE                   \\ 
\midrule
\multirow{4}{*}{ETTh1} & 96   & 0.056 & 0.181                  & 0.054          & 0.178          & \uline{0.054}  & \textbf{0.177} & \uline{0.054}           & 0.181          & 0.189 & 0.359               & \textbf{0.053} & \textbf{0.177} & 0.056 & 0.180               & 0.079                    & 0.215 & 0.080          & 0.214           & 0.071         & 0.206           & 0.193 & 0.377                 & 0.283 & 0.468                 \\
                       & 192  & 0.071 & \uline{0.205}          & \textbf{0.069} & \uline{0.205}  & 0.071          & 0.206          & 0.073                   & 0.214          & 0.078 & 0.212               & \textbf{0.069} & \textbf{0.204} & 0.071 & 0.204               & 0.104                    & 0.245 & 0.105          & 0.256           & 0.114         & 0.262           & 0.217 & 0.395                 & 0.234 & 0.409                 \\
\bottomrule
\end{tabular}}}
\label{sup_table:Uni-data-results}
\end{table*}

\end{document}